\documentclass[10pt]{iopart}
\expandafter\let\csname equation*\endcsname\relax
\expandafter\let\csname endequation*\endcsname\relax

\usepackage{adjustbox,xspace}
\usepackage[hang,flushmargin]{footmisc}
\usepackage{siunitx}
\usepackage{mathtools,gensymb,xfrac,multicol,multirow,resizegather,bm}  
\usepackage[standard]{ntheorem}  
\usepackage{booktabs,xcolor,enumitem,paralist} 
\usepackage{floatrow, subcaption,graphicx}    
\usepackage{colortbl,tabularx}
\usepackage{dblfloatfix} 

\usepackage{relsize} 

\usepackage{floatrow}

\DeclareTextFontCommand{\textbfit}{%
  \fontseries\bfdefault 
  \itshape
}

\theoremseparator{.}
\theorembodyfont{\upshape}
\newtheorem{hyp}{Hypothesis}
\makeatletter
\newcounter{subhyp} 
\let\savedc@hyp\c@hyp

\newcommand{\normhyp}{%
  \let\c@hyp\savedc@hyp 
  \renewcommand\thehyp{\arabic{hyp}}%
} 
\makeatother

\theoremseparator{.}
\theorembodyfont{\upshape}
\newtheorem{contr}{Contribution}
\makeatletter
\newcounter{subcontr} 
\let\savedc@contr\c@contr

\newcommand{\normcontr}{%
  \let\c@contr\savedc@contr 
  \renewcommand\thecontr{\arabic{contr}}%
} 
\makeatother

\makeatletter
\newcounter{subclaw} 
\let\savedc@claw\c@claw
\makeatother

\newtheorem{event}{Condition}
\makeatletter
\newcounter{subevent} 
\let\savedc@event\c@event
\makeatother

\newcommand{\minus}{\scalebox{0.75}[1.0]{$-$}}    
\newcommand{\mminus}{\scalebox{0.65}[1.0]{$-$}}   
\newcommand{\mytilde}{\raise.17ex\hbox{$\scriptstyle\mathtt{\sim}$}} 
\newcommand\myeq{\mkern1.5mu{=}\mkern1.5mu}
\newcommand\mycross{\mkern1.5mu{\times}\mkern1.5mu}
\newcommand\mygeq{\mkern1.5mu{\geq}\mkern1.5mu}

\newcommand\myleq{\mkern1.5mu{\leq}\mkern1.5mu}

\newcommand{\VP}{\textrm{VP}\xspace}

\newcommand{\GRF}{\textrm{GRF}\xspace}

\newcommand{\CoM}{\textrm{CoM}\xspace}
\newcommand{\TSLIP}{\textrm{TSLIP}\xspace}

\newcommand\VPa[1][color_VPBlue]{$\mathrm{VP_{\color{#1}{A}}}$\xspace}
\newcommand\VPb[1][color_VPRose]{$\mathrm{VP_{\color{#1}{B}}}$\xspace} 
\newcommand\VPbl[1][color_VPRoseDark]{$\mathrm{VP_{\color{#1}{BL}}}$\xspace} 

\newcommand{\mcref}[1]{{Figure\,{\labelcref{#1}}}} 
\newcommand{\multiref}[3]{Figure\,{\labelcref{#1}{#2}}~and~{\labelcref{#1}{#3}}} 
\newcommand{\multirefcomma}[3]{Figure\,{\labelcref{#1}{#2}},{\labelcref{#1}{#3}}} 
\newcommand{\multirefrange}[3]{Figure\,{\labelcref{#1}{#2}}-{\labelcref{#1}{#3}}} 
\newcommand{\multirefrangewo}[3]{{\labelcref{#1}{#2}}-{\labelcref{#1}{#3}}} 
\usepackage[compact]{titlesec}

\usepackage{tikz,scalerel}
\usetikzlibrary{svg.path}
\definecolor{orcidlogocol}{HTML}{A6CE39}
\tikzset{
  orcidlogo/.pic={
    \fill[orcidlogocol] svg{M256,128c0,70.7-57.3,128-128,128C57.3,256,0,198.7,0,128C0,57.3,57.3,0,128,0C198.7,0,256,57.3,256,128z};
    \fill[white] svg{M86.3,186.2H70.9V79.1h15.4v48.4V186.2z}
                 svg{M108.9,79.1h41.6c39.6,0,57,28.3,57,53.6c0,27.5-21.5,53.6-56.8,53.6h-41.8V79.1z M124.3,172.4h24.5c34.9,0,42.9-26.5,42.9-39.7c0-21.5-13.7-39.7-43.7-39.7h-23.7V172.4z}
                 svg{M88.7,56.8c0,5.5-4.5,10.1-10.1,10.1c-5.6,0-10.1-4.6-10.1-10.1c0-5.6,4.5-10.1,10.1-10.1C84.2,46.7,88.7,51.3,88.7,56.8z};
  }
}
\newcommand\orcidicon[1]{\href{https://orcid.org/#1}{\mbox{\scalerel*{
\begin{tikzpicture}[yscale=-1,transform shape]
\pic{orcidlogo};
\end{tikzpicture}
}{|}}}}
       
\usepackage{hyperref} 

\usepackage[noadjust]{cite} 

\usepackage[capitalize,noabbrev,nameinlink]{cleveref} 
\newcommand{\cmmnt}[1]{\ignorespaces}

\crefname{hyp}{Hyp.}{Hyp.}
\crefname{claw}{Rule}{Rule}
\crefname{event}{Cond.}{Cond.}

\usepackage{etoolbox}
\makeatletter
\newcommand{\mainmatter}{%
  \setcounter{footnote}{0}%
  \let\@fnsymbol\@arabic
  \def\@makefnmark{\textsuperscript{\arabic{footnote}}}%
}
\makeatother

\usepackage{footnotehyper}

\usepackage{tikz}
\usetikzlibrary{shapes, calc, arrows, shadows, positioning, bending,tikzmark,decorations.pathreplacing}

\newenvironment {annotatedFigure}[1]{\centering\begin{tikzpicture}[remember picture]
\node[anchor=south west,inner sep=0] (image) at (0,0) {#1};\begin{scope}[x={(image.south east)},y={(image.north west)}]}{\end{scope}\end{tikzpicture}}

\newcommand*\customSublabelBasic[3]{
\node at (#2) [fill=none,shape=circle,draw=none, scale=1,inner sep=1pt,font=\sffamily,text=#3] {\text{\footnotesize #1}};}

\tikzset{
  do path picture/.style={%
    path picture={%
      \pgfpointdiff{\pgfpointanchor{path picture bounding box}{south west}}%
        {\pgfpointanchor{path picture bounding box}{north east}}%
      \pgfgetlastxy\x\y%
      \tikzset{x=\x/2,y=\y/2}%
      #1
    }
  },
  sinwave/.style={do path picture={    
    \draw [line cap=round] (-3/4,0)
      sin (-3/8,1/2) cos (0,0) sin (3/8,-1/2) cos (3/4,0);
  }},
  coswave/.style ={do path picture={    
    \draw [line cap=round] (-3/4,0)
      sin (-3/8,-1/2) cos (0,0) sin (3/8,1/2) cos (3/4,0);
  }},
  circlecross/.style={do path picture={    
    \draw [line cap=round,black] (-0.6,-0.6) -- (0.6,0.6) (-0.6,0.6) -- (0.6,-0.6);
  }},
  plus/.style={do path picture={    
    \draw [line cap=round] (-3/4,0) -- (3/4,0) (0,-3/4) -- (0,3/4);
  }}
}
\newcommand{\markerSineA}{\raisebox{-1 pt}{\tikz{\node[circle,draw={color_VPBlue},coswave,scale=1,opacity=1](){};}}}
\newcommand{\markerSineB}{\raisebox{-1 pt}{\tikz{\node[circle,draw={color_VPRose},scale=1,sinwave,fill=white,opacity=1](){};}}}
\usepackage{amssymb}
\newcommand\circledcheckmark[1][color_VPcom]{\raisebox{-1 pt}
{\tikz\node[draw=#1,scale=0.8,circle,fill=none,inner sep=0pt]{$\checkmark$};}}
\newcommand\circledcross[1][color_VPcom]{\raisebox{-1 pt}
 {\tikz\node[draw=#1,scale=0.9,circle,circlecross,fill=none]{};}}

\newcommand\markerVPa[1][color_VPBlue]{\raisebox{-0.5 pt}{\tikz{\node[draw=white,scale=0.6,circle,fill=#1,opacity=1](){};}}}
\newcommand\markerVPb[1][color_VPRose]{\raisebox{-0.5 pt}{\tikz{\node[draw=white,scale=0.6,circle,fill=#1,opacity=1](){};}}}
\newcommand\markerVPcom[1][color_VPcom]{\raisebox{-0.5 pt}{\tikz{\node[draw=white,scale=0.6,shape=circle,fill=#1,opacity=1](){};}}\xspace}

\newcommand{\markerLit}{\raisebox{1 pt}{\tikz{\draw[line width=0.5mm, {color_Gfowl3}](0,0)--(0.3,0);}}}

\tikzset{cross/.style={cross out, draw, fill=none, minimum size=2*(#1-\pgflinewidth), inner sep=0pt, outer sep=0pt}, cross/.default={3.5pt}}
\newcommand\markerDamperData[1][color_DamperData]{\raisebox{-0.2 pt}{\tikz{\node[draw={#1},scale=1,very thick,cross,rotate=90](){};}}}
\newcommand\markerGfowl[1][color_Gfowl2]{\raisebox{-0.2 pt}{\tikz{\node[draw={#1},scale=1,very thick,cross,rotate=90](){};}}}
\newcommand\markerGfowlR[1][color_Gfowl1]{\raisebox{-0.2 pt}{\tikz{\node[draw={#1},scale=1,very thick,cross,rotate=90](){};}}}

\newcommand\markerCrossGreen[1][color_Biliniar]{\raisebox{-0.2 pt}{\tikz{\node[draw={#1},scale=1,very thick,cross,rotate=90](){};}}}
\newcommand\markerLineGrey[1][color_mygray]{\raisebox{1 pt}{\tikz{\draw[line width=0.6mm, draw={#1}](0,0)--(0.3,0);}}}

\newcommand\mylabelA[1][color_mygray]{\raisebox{0.5 pt}{\footnotesize{\textbf{\textcolor{#1}{(A)}}}}}
\newcommand\mylabelB[1][color_mygray]{\raisebox{0.5 pt}{\footnotesize{\textbf{\textcolor{#1}{(B)}}}}}
\newcommand\mylabelC[1][color_mygray]{\raisebox{0.5 pt}{\footnotesize{\textbf{\textcolor{#1}{(C)}}}}}
\newcommand\mylabelD[1][color_mygray]{\raisebox{0.5 pt}{\footnotesize{\textbf{\textcolor{#1}{(D)}}}}}

 \newcommand\markerDmpLin[1][color_Linear]{\raisebox{-1 pt}{\tikz{\node[draw=white, isosceles triangle, isosceles triangle apex angle=60, inner sep=0pt, anchor=lower side, rotate=90, line width=1.5pt, minimum height=0.27cm, fill={#1}] (triangle) at (0,0) {};}}\xspace}

\newcommand\markerDmpBiLin[1][color_Biliniar]{\raisebox{-0.5 pt}{\tikz{\node[draw=white,scale=0.8,shape=rectangle,fill={#1}](){};}}\xspace}

\newcommand{\arrowWorkP}{\raisebox{-2 pt}{\tikz{\draw[->, dash pattern={on 3pt off 1 pt on 1pt off 1pt on 3pt off 1 pt on 1pt off 1pt}, line width=0.35mm, {color_Wp}](0, 0.01)--(0, 0.35);}}}
\newcommand{\arrowWorkN}{\raisebox{-2 pt}{\tikz{\draw[->, dash pattern={on 3pt off 1 pt on 1pt off 1pt on 3pt off 1 pt on 1pt off 1pt}, line width=0.35mm, {color_Wn}](0, 0.35)--(0, 0.01);}}}

\newcommand{\arrowShort}{\raisebox{1.5 pt}{\tikz{\draw[->,line width=0.35mm, {color_myblack}](0.01,0)--(0.19,0);}}}

\newcommand\arrowPink[1][color_myblack]{\raisebox{1.5 pt}{\tikz{\draw[<-,line width=0.45mm, draw={#1}](0.01,0)--(0.4,0);}}}

\newcommand\arrowAcc[1][color_GRFacc]{\raisebox{1.5 pt}{\tikz{\draw[->,line width=0.35mm, draw={#1}](0.01,0)--(0.3,0);}}}
\newcommand\arrowDec[1][color_GRFdec]{\raisebox{1.5 pt}{\tikz{\draw[<-,line width=0.35mm, draw={#1}](0.01,0)--(0.3,0);}}}

\newcommand\arrowCircCw[1][color_mygray]{\raisebox{-2 pt}{\tikz{\draw[thick, -{>[flex=0.75]}, line width=0.35mm, draw={#1}] (0,0) arc[radius=0.15cm,start angle=220,delta angle=330];}}}
\newcommand\arrowCircCcw[1][color_mygray]{\raisebox{-2 pt}{\tikz{\draw[thick, {<[flex=0.75]}-, line width=0.35mm, draw={#1}] (0,0) arc[radius=0.15cm,start angle=0,delta angle=330];}}}

\begin{document}

\makeatletter  
\def\mathcolor#1#{\@mathcolor{#1}}
\def\@mathcolor#1#2#3{%
  \protect\leavevmode
  \begingroup
    \color#1{#2}#3%
  \endgroup
}
\makeatother 

\definecolor{color_VPRose}{rgb}{0.6980    0.0941    0.1686} 
\definecolor{color_VPRoseDark}{rgb}{0.3017    0.0004    0.0026} 
\definecolor{color_VPBlue}{rgb}{0.1294    0.4000    0.6745}
\definecolor{color_VPBlueDark}{rgb}{0.0011    0.0472    0.2691} %

\definecolor{color_VPb_B}{rgb}{ 0.8823    0.4977    0.4251}
\definecolor{color_VPb_A}{rgb}{0.9488    0.7460    0.6982}
\definecolor{color_VPcom}{rgb}{0.5 0.5 0.5}
\definecolor{color_VPver}{rgb}{0.2235    0.0667    0.3098}


\definecolor{color_GRFacc}{rgb}{0.0745    0.7098    0.3255}
\definecolor{color_GRFdec}{rgb}{0.8627         0    0.4706}
\definecolor{color_myblack}{rgb}{0.1 0.1 0.1}
\definecolor{color_mygray}{rgb}{0.5 0.5 0.5}

\definecolor{color_Wp}{rgb}{0.584 0.643 0.737} 
\definecolor{color_Wn}{rgb}{0.666  0.596 0.611} 

\definecolor{color_CW}{rgb}{0.5020    0.8196    0.8510}
\definecolor{color_CCW}{rgb}{0.9922    0.6314    0.4392}

\definecolor{color_PowderBlue}{rgb}{0.7176    0.8706    0.8431}
\definecolor{color_Thistle}{rgb}{0.8078    0.7373    0.8667}
 \definecolor{color_Seagreen}{rgb}{0.3528    0.5041    0.4233}
 \definecolor{color_BlueGrey}{rgb}{0.4 0.4 0.4}
 
\definecolor{color_Gfowl}{rgb}{0.3451    0.1647    0.6980}
\definecolor{color_Gfowl1}{rgb}{0.7037    0.5748    0.8191}
\definecolor{color_Gfowl2}{rgb}{0.8389    0.7582    0.9050}
\definecolor{color_Gfowl3}{rgb}{ 0.1725    0.0627    0.3686}

\definecolor{color_DamperData}{rgb}{0.3451    0.1647    0.6980}
\definecolor{color_Biliniar}{rgb}{0    0.4275    0.1725}
\definecolor{color_Linear}{rgb}{0.5804    0.8275    0.7765}

\definecolor{color_B_VPa}{rgb}{0.2627    0.7765    0.8353}
\definecolor{color_W_VPa}{rgb}{0.1843    0.5922    0.8353}
\definecolor{color_B_VPb}{rgb}{ 1.0000    0.5647    0.3333}
\definecolor{color_W_VPb}{rgb}{1.0000    0.3216    0.3333}

\title[]{Trunk Pitch Oscillations for Energy Trade-offs in Bipedal Running Birds and Robots}
\author{{\"O}zge Drama \orcidicon{0000-0001-7752-0950}, Alexander Badri-Sp{\"o}witz \orcidicon{0000-0002-3864-7307}}
\address{Dynamic Locomotion Group, Max Planck Institute of Intelligent Systems, Germany} 
\ead{drama@is.mpg.de}
\vspace{10pt}

\begin{indented}
\item[]September 2019
\end{indented}

\begin{abstract}
Bipedal animals have diverse morphologies and advanced locomotion abilities. Terrestrial birds, in particular, display agile, efficient, and robust running motion, in which they exploit the interplay between the body segment masses and moment of inertias.
%
On the other hand, most legged robots are not able to generate such versatile and energy-efficient motion and often disregard trunk movements as a means to enhance their locomotion capabilities. 
%
Recent research investigated how trunk motions affect the gait characteristics of humans, but there is a lack of analysis across different bipedal morphologies.
%
To address this issue, we analyze avian running based on a spring-loaded inverted pendulum model with a pronograde (horizontal) trunk.
%
We use a virtual point based control scheme and modify the alignment of the ground reaction forces to assess how our control strategy influences the trunk pitch oscillations and energetics of the locomotion.
%
We derive three potential key strategies to leverage trunk pitch motions that minimize either the energy fluctuations of the center of mass or the work performed by the hip and leg. We suggest how these strategies could be used in legged robotics. 
\end{abstract}
%
%
{\submitto{\BB}}
%
%
%
%
\ioptwocol
%
%
\mainmatter
%
%
\section{Introduction}\label{sec:Introduction}
Creating dynamic running motion for bipedal mechanisms is challenging due to the complexity in controlling an underactuated trunk-leg structure, that has nonlinear coupled dynamics and intermittent ground contacts~\cite{Sharbafi_2017_Book}.~Robotics research often focuses on the control of lower extremities for humanoids and suppresses trunk motions for simplicity ~\cite{Raibert_1986,Westervelt_2007}.~In contrast, bipedal animals have diverse morphologies and display a wide range of motion patterns with prominent trunk movements~\cite{Hutchinson_2004_I}.~In particular, terrestrial birds are able to generate exceptionally agile, energy efficient, and robust motion; irrespective of their vast variability in body size, posture, and habitat~\cite{Clemente_2018,Daley_2018,Hutchinson_2004_I}. Birds exploit their trunk's inertia and generate trunk movements to assist the postural stability~\cite{Abourachid_2011,Hancock_2014}.~The concept of leveraging trunk pitch oscillations has been analyzed for humanoids, where the trunk motion assists energetics by redistributing the work between the leg and hip joint~\cite{Drama_2019}. Here, we investigate whether a similar strategy exists for terrestrial birds with a pronograde (horizontal) trunk orientation. We use a spring-loaded inverted pendulum model with a controller based on a virtual point (\VP) concept. Our aim is to analyze how the magnitude and direction of the oscillations depend on the \VP position and running speed.  

Birds show exceptional locomotor efficiency and capabilities, yet the mechanisms underlying such agile motion are not well understood \cite{Daley_2018}. The majority of the prior research builds on simplified models that address the motion of the center of mass (\CoM) without the trunk \cite{BirnJeffery_2014,Blum_2011,Daley_2018}. These studies are able to recreate some of the essential characteristics of avian gaits. In particular, these gaits display asymmetric traits such as left-skewed ground reaction forces (\GRF)~\cite{Andrada_2014_A,BirnJeffery_2014,Clemente_2018}, distinct leg lengths at leg touch-down/take-off~\cite{Andrada_2014_A,BirnJeffery_2014}, and asymmetric leg protraction/retraction angles \cite{Gatesy_1991, Gatesy_1999, Abourachid_2000_R}. Recent studies reveal the role of the pronograde (horizontal) trunk orientation in these gait asymmetries~\cite{Aminiaghdam_2017,Andrada_2014,Andrada_2015,Blickhan_2015}. Terrestrial birds show pronograde trunk posture, where the \CoM is located cranial (headward) to the hip joint with an inclination of \SIrange[range-phrase=-]{100}{135}{\degree} with respect to the vertical axis \cite{Hutchinson_2004_I}. Bird's unique hip attachment necessitates continuous hip extension torques to hold the trunk in a horizontal orientation against gravity \cite{Andrada_2014}. Furthermore, the trunk amounts to \SIrange[range-phrase=-,range-units=single]{70}{80}{\percent} of the total body mass and has a high moment of inertia \cite{Abourachid_2000_R,Fedak_1982,Hutchinson_2004_I}. The heavy, horizontal trunk requires high hip torques, which can be realized only if the inertia is high enough to resist the trunk rotation. Hence, the trunk is an integral part of understanding bipedal locomotion.

Experimental data related to the avian trunk motion is available only sparsely. Studies show~that the avian trunk pitches downward (ventrally) during the double stance phases of walking \cite{Abourachid_2011,Hancock_2014,Hugel_2011},~and breaking phase of running \cite{Gatesy_1999,Hancock_2014,Jindrich_2007}.~The trunk moves upward (dorsally) for the remaining phases of the gait.~Experiments report \SI{4}{\degree}~trunk angular excursion ($\Delta \theta$) for elegant crested tinamous moving at \SI{1.74}{\metre\per\second}~\cite{Hancock_2014},~$\Delta \theta \myleq \SI{10}{\degree}$~for guineafowls/quails \cite{Abourachid_2011,Gatesy_1999},~and $\Delta \theta \myleq \SI{6}{\degree}$ for ostriches moving at \SI{3.3}{\metre\per\second} \cite{Jindrich_2007,Rubenson_2007}. It is not yet understood what effect trunk motions have on the gait.

The spring-loaded inverted pendulum (SLIP) model can predict fundamental characteristics of running motion for animals with varying morphologies. Its simplicity and its prediction capabilities make the model well suited to investigate both humans and avians \cite{Blickhan_1989}. The model consists of a point-mass body attached to a massless and springy leg. The SLIP can be extended with a rigid trunk (\TSLIP), which is actuated by a torque at the hip \cite{Andrada_2014}. One method to select the hip torque is based on the \VP concept, where the ground reaction forces are redirected to intersect at a virtual point. The \GRF intersecting at a point bounds the moment \GRF creates around the \CoM. Therefore, some research considers the \VP as a pivot point that assists postural stability, and implement the \VP as a control target in both avian and human \TSLIP models \cite{Andrada_2014,Maus_2010,Sharbafi_2017_Book}.

The \VP can also be regarded as a method to generate trunk oscillations with different magnitudes and directions, which is shown in~\cite{Drama_2019} for the human morphology. The relative position of the \VP with respect to the axis passing through the \CoM and foot determines the direction of the trunk rotation. The trunk rotates backward during the stance phase if the the \VP is {\color{color_VPBlue} above} (\VPa) the \CoM-foot axis, whereas it rotates forward if \VP is {\color{color_VPRose} below} (\VPb)~\cite{Drama_2019}. The distance between the \VP~and~\CoM determines the magnitude of the trunk oscillations. As a consequence of the direct mapping between \VP location and trunk oscillations, there is no need to adjust the control gains when the operating point of the gait changes. Hence,~the framework allows creating a variety of trunk oscillations across a range of speeds in a systematic and comparable manner. 

Another advantage of the \VP framework is that it generates feasible hip torque profiles. The \VP concept creates a coupling between the magnitudes of the hip torque and leg force. Consequently, the hip torque profile inherits the continuity and smoothness properties of the leg force profile. It makes the \VP framework well suited for investigating the role of the hip torque in generating trunk motions.

The prospect of using the \VP framework for the pronograde trunk is motivated by the avian gait measurements. A \VPa is observed in the running gait of chickens \cite{Maus_2010}, and in walking and running gait of quails \cite{Andrada_2014, Blickhan_2015}. Inspired by these observations, the \VP is implemented for the pronograde \TSLIP model \cite{Andrada_2014}. What is missing is the relation of the \VP to the trunk oscillations for the avian morphology. It is shown that a \VPa in the \TSLIP framework produces backward trunk motion, which is the opposite of what is observed in human running \cite{Drama_2019}. Instead, a \VPb provides the matching forwards trunk rotation in the human model. Here, we inquire what relation exists for the avian morphology.

Our overall goal is to conceptualize how trunk motion is beneficial for bipeds with different trunk orientations. We would like to transfer the gained knowledge to controller and robot design.Currently, robot studies often maintain the trunk posture at a fixed angle without any movements. There are only a few studies that address dynamic trunk stability; one of which uses a \VP control scheme \cite{Peekema_2015} for walking gaits, while the other tracks a sinusoidal reference pitch angle \cite{Rezazadeh_2015} within the \TSLIP framework for the ATRIAS robot. However, these studies involve a single parameter set and do not extend to multiple operating points with multiple speeds and different trunk oscillation characteristics.

In this paper, we present a unified framework to analyze the effect of trunk motions for bipedal running, with a focus on avians. We implement a \TSLIP model with varying \VP targets for a pronograde trunk and systematically compare the resulting gaits for speeds of \SIrange[range-phrase=-,range-units=single]{4}{10}{\metre\per\second}. 
%
Our work examines four hypotheses, where we question if the pronograde \TSLIP model is able to 
\begin{compactitem}[]
\setlength{\itemsep}{0.1cm}
\item \begin{hyp}\label{hyp:1_GRF}
 predict left-skewed \GRF profiles, whose magnitude is proportional to the forward speed and matches to the avian gait data in~\cite{Bishop_2017,Clemente_2018,Gatesy_1991}.
\end{hyp} 
\item  \begin{hyp}\label{hyp:2_thB}
 generate downward (dorsoventral) trunk pitch motion at stance phase, whose magnitude is proportional to the forward speed \cite{Drama_2019}.
\end{hyp}
%
\item  \begin{hyp}\label{hyp:3_VP}
 utilize the \VP to alter the gait dynamics, which gives rise to multiple solutions with different gait characteristics for a given speed.
\end{hyp}
\item \begin{hyp}\label{hyp:4_VPGRF}
 determine the \VP location in favor of energetics, in a similar manner as \cite{Drama_2019}.
\end{hyp}
\end{compactitem}

This work presents three novel contributions that are essential for the pronograde \TSLIP model.
\begin{compactitem}[]
\setlength{\itemsep}{0.1cm}
\item \begin{contr}\label{contr:1_Bilin_c}
We show that a bilinear leg damper \cite{Abraham_2015} generates zero damping force at leg touch-down, and therefore is better suited for the pronograde \TSLIP model compared to a typical linear leg damper \cite{Andrada_2014,Sharbafi_2013}. 
\end{contr}
\item \begin{contr}\label{contr:2_Frames}
We investigate which frame is suitable for defining the \VP location to obtain feasible gaits. The \VP is typically defined with respect to the body frame and above~the \CoM in the \TSLIP model to facilitate a self-stabilizing postural behavior \cite{Maus_2010}. There is little evidence in literature to support the choice of the \VP frame. It is shown that a body or world aligned \VP predicts the \GRF better than a trunk aligned \VP in human walking \cite{Gruben_2012}. We extend this analysis to avian running using the \TSLIP model.
\end{contr}
\item \begin{contr}\label{contr:3_VP_GRF}
The change in \VP position corresponds to modifying the orientation of the \GRF vector \cite{Drama_2019}. A \VPa creates more vertically oriented \GRF compared to \VPb. A more vertically oriented~\GRF vector has a smaller horizontal GRF component, which accelerates and decelerates the body less (i.e., it is energetically more favorable)~\cite{Clark_1975,Usherwood_2012}.
In a similar manner, we test whether a \VPa in our avian \TSLIP model can yield such an energetic benefit. In particular, we establish a relation between the \VP location, \GRF alignment and energy variations of the \CoM.
\end{contr}
\end{compactitem}
%

\section{Simulation Model}\label{sec:SimModel}
\begin{figure}[!t]
\centering
\begin{annotatedFigure}
	{\includegraphics[width=1.0\columnwidth]{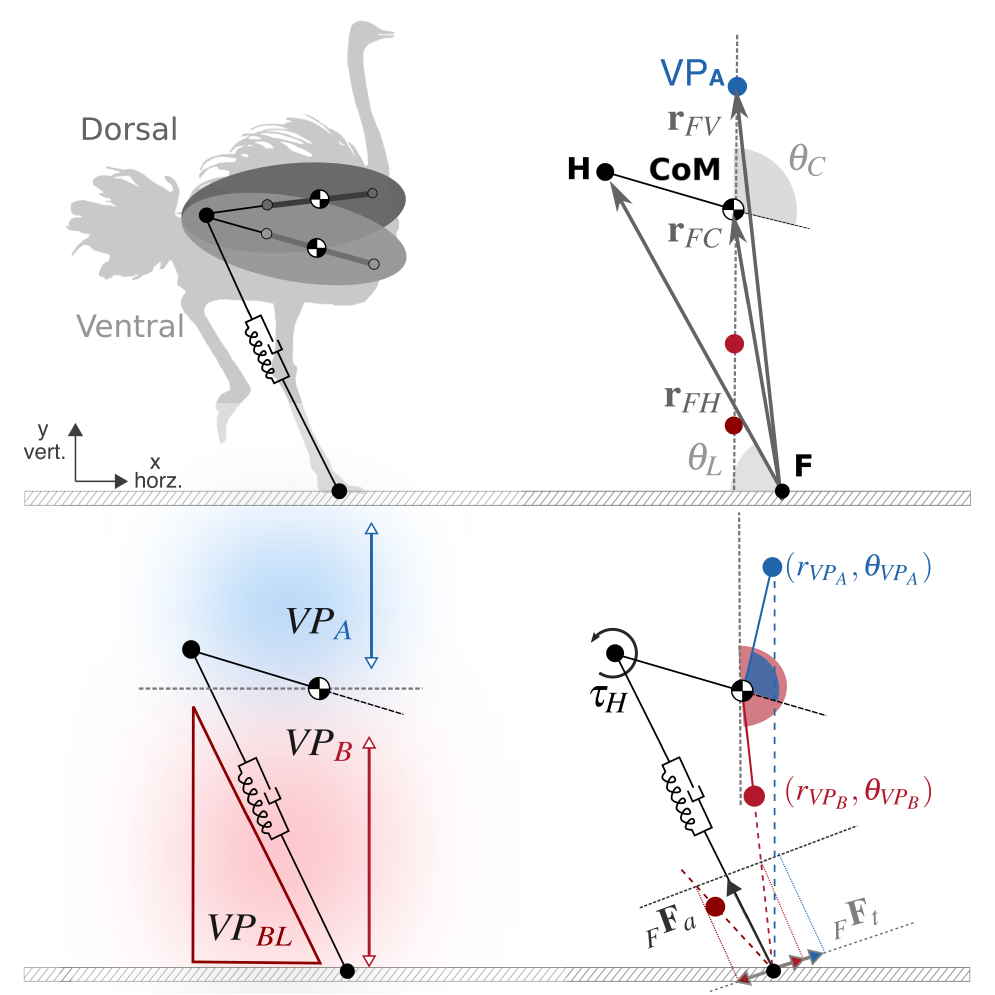}}
	\customSublabelBasic{a)}{0.05,0.95}{color_mygray}
	\customSublabelBasic{c)}{0.05,0.42}{color_mygray}
	\customSublabelBasic{b)}{0.535,0.95}{color_mygray}
	\customSublabelBasic{d)}{0.535,0.42}{color_mygray}
\end{annotatedFigure}
\captionof{figure}{a) Avian \TSLIP model with a pronograde (horizontal) trunk in dorsal (upward) and ventral (downward) positions{\protect\footnotemark}. b) Vector notations for the \TSLIP model. Letters $\mathrm{VP, \:  CoM, \: H}$ denote the virtual point, center of mass, and hip, respectively. Position vectors are referred to as $\mathbf{r}_{FH}, \mathbf{r}_{FV}, \mathbf{r}_{FC}, \mathbf{r}_{FH}$. Angles $\mathbf{\theta}_{L}, \mathbf{\theta}_{C},\mathbf{\theta}_{VP}$ are the leg, trunk and \VP angles. c) Classification of regions for placing the virtual point. There are two major axes that affects the trunk motion: \CoM-foot and hip-foot axes. The \CoM-foot axis sets the direction of the angular moment applied by the \GRF. Virtual points above the \CoM-foot axis cause mainly upward trunk motion during stance, whereas points below yield the opposite. \VP needs to lie on the vertical axis passing through the \CoM for the steady state motion~\cite{Sharbafi_2013}. Hence, the parameter space is divided between the \VP's above the \CoM (\VPa) and below the \CoM (\VPb). The hip-foot axis determines the sign of the hip torque.  All \VPa lie above hip-foot axis for our parameter range (\VP radius up to \SI{60}{\centi\meter}). The \VPb region is divided into points above the hip-foot axis, and below. We refer the points below both \CoM-foot and hip-foot axes as \VPbl. d) The leg spring and damper create the axial component of the ground reaction force $ \prescript{}{F}{\mathbf{F}}_{a}$, whereas the hip torque creates the tangential component $ \prescript{}{F}{\mathbf{F}}_{t}$. The vector sum of these forces passes through the virtual point VP.}
 \label{fig:TSLIP}
\footnotetext{\, The ostrich silhouette is adapted with the permission from \text{https://proartinc.net/shop/4k-wildlife-relax/4k-ostrich-africa}.}
\end{figure}

In this section, we describe the TSLIP model applied in this work. It consists of a trunk with mass $m$ and moment of inertia $J$, which is connected to a massless leg of length $l$ that has a parallel spring-damper mechanism (see \mcref{fig:TSLIP}a). The dynamics are described by a flight phase, where the \CoM moves in a ballistic motion and a stance phase, where the leg force and hip torque propel the body forward. The switch between these phases occurs at the touch-down (TD) and take-off (TO) points, where the foot establishes contact with the ground and the leg extends to its rest length $l_{0}$, respectively. Swing leg dynamics are omitted, similar to other TSLIP studies in \cite{Andrada_2014,Maus_2008}.

The equations of motion for the \CoM state $(x_{C},  y_{C},  \theta_{C}) $ during the stance phase can be written~as, 
\begin{equation}
\begin{aligned}
m \begin{bmatrix}  \ddot{x}_{C} \\  \ddot{y}_{C} \end{bmatrix} & \myeq \prescript{}{F}{\mathbf{F}}_{a} +  \prescript{}{F}{\mathbf{F}}_{t} + g, \\
J\, \ddot{\theta}_{C} & \myeq \minus {\mathbf{r}}_{FC} \mycross (  \prescript{}{F}{\mathbf{F}}_{a} +  \prescript{}{F}{\mathbf{F}}_{t}).
\label{eqn:EoM} 
\end{aligned}
\end{equation}
The linear leg spring force $F_{sp} \myeq k \, (l \minus l_{0})$ and bilinear leg damping force $F_{dp} \myeq c \, \dot{l} \, (l  \minus l_{0})$ generate the axial component of the \GRF in foot frame $ \prescript{}{F}{\mathbf{F}}_{a}  \myeq \left(F_{sp} \minus F_{dp} \right)  
 {\left[ \mminus \cos\theta_{L} \; \sin\theta_{L}  \right]}^\mathsf{T}$. The hip torque ${\tau}_{H}$  generates the tangential component of the \GRF $ \prescript{}{F}{\mathbf{F}}_{t} \myeq \left( \scaleto{\sfrac{\minus \tau_{H}}{l_{L}}}{12pt} \right)  
 {\left[ \sin\theta_{L} \;  \mminus \cos\theta_{L}  \right] }^\mathsf{T},$ as seen in \mcref{fig:TSLIP}d. In our formulation, $k$ denotes the stiffness of the leg spring and $c$ is the damping coefficient.
 
The leg is passively compliant, where the damper removes energy from the system by performing negative work, and the spring stores and releases elastic energy in sequence. The hip is actuated and produces \emph{net} positive work to balance the energy depleted by the leg.
%
Thus, the actuated hip torque $\tau_{H}$ is the only element that we can actively control to induce trunk pitch oscillations. We select $\tau_{H}$, such that the \GRF points to a \VP, which is characterized by the radius $r_{VP}$ (i.e.,~distance between the hip and \CoM) and angle $\theta_{VP}$, as shown in \mcref{fig:TSLIP}d ({\protect \markerVPa,\protect \markerVPb }). The hip torque as a function of the VP is expressed as,
%
\begin{equation}
\begin{aligned}
\tau_{H} &= \tau_{VP} =  \prescript{}{F}{\mathbf{F}}_{a} \times  \left[   \frac{\mathbf{r}_{FV} \times \mathbf{r}_{FH} }{\mathbf{r}_{FV} \cdot \mathbf{r}_{FH}}\right]   \times  l,  \\
\mathbf{r}_{FV} &= \mathbf{r}_{FC}  + r_{VP}    \begin{bmatrix*}[r] \minus \sin \left( \theta_{C}+\theta_{VP} \right) \\ \cos \left( \theta_{C}+\theta_{VP} \right) \end{bmatrix*}.
\end{aligned}
\label{eqn:tauVP}
\end{equation}
%
\indent The sign of the hip torque depends on whether the VP is placed above or below the the hip-foot axis (i.e., leg axis). For the pronograde trunk, all \VPa within our parameter range remain above the leg axis across entire stance phase. Thus, the resulting hip torque is always negative. \VPb above the leg axis yield similar torque pattern as \VPa. In contrast, \VPb below the leg axis yield positive hip torque at the first half of the stance phase and negative towards the second half. These region is referred to as  \VPbl, and is illustrated in \mcref{fig:TSLIP}c.

The concept of \VP control is open-loop, therefore the \VP controller is highly sensitive to changes in the initial state and model parameters. This parameter sensitivity makes it challenging to find feasible gaits. In order to simplify and guide the parameter search, we use the iterative gait generation framework in \cite{Drama_2019}. The framework has an initial controller that combines the \VP based torque in \cref{eqn:tauVP} with a PID controller on the pitch angle, which yields stable gaits with semi-focused \GRF in \multiref{fig:VP}{b}{c}.~The solutions are fed to an intermediate control scheme, in which the PID control is disabled and the \VP angle is linearly adjusted with respect to the body angle. With this scheme, simulated gaits converge to a \VP trend, where the \GRF gradually focus on a single point as in \multiref{fig:VP}{d}{e}. 

The morphological parameters of the \TSLIP model are selected from the biomechanical literature to match an ostrich of \SI{80}{\kilogram} with \SI{1}{\meter} leg length (see \cref{tab:ModelPrm}). The control gains and damping coefficient of the model are then adjusted to conform certain gait characteristics observed in avian locomotion, which is described in detail in \cref{subsec:ControlMod}~and~\labelcref{subsec:ExpSetup}. 

\begin{table}[h!]                                                                                                                                                                                                                                                                                                                                                                                                                                                                                                                                                                 
\centering                                                                                                                                                                                                                                                                                                                                                                                                                                                                                                                                                                          
\captionsetup{justification=centering}
\caption{Model parameters for avian{\protect\footnotemark} TSLIP model. Damping parameters are iteratively set  in \mcref{fig:Gaits_Damping_th0}a.}
\label{tab:ModelPrm}
\begin{adjustbox}{width=0.97\textwidth}
\begin{tabular}{@{} l| c c c c c  @{}}
\multicolumn{1}{l}{Name} & \multicolumn{1}{c}{Symbol} &  \multicolumn{1}{c}{Units} &   \multicolumn{1}{c}{Literature} & \multicolumn{1}{c}{Chosen} & \multicolumn{1}{c}{Reference}  \\
\hline
\hspace{1mm} mass & $\mathit{m}$  & \si{\kilogram}    &  70-100  & 80 &  {\cite{Abourachid_2000_R,Fedak_1982}}   \\
\hspace{1mm} moment of inertia & $\mathit{J}$   & \si{\kilogram\per\meter\squared} & 10  & 10 &  {\cite{Fedak_1982,Hutchinson_2004_I}}    \\ 
\hspace{1mm} trunk angle & $\theta_{C}$ & (\si{\degree}) & 93-125{\protect\footnotemark}  & 100 & \cite{Hutchinson_2004_I,Jindrich_2007,Rubenson_2007} \\
\hspace{1mm} leg stiffness & $\mathit{k}$  & \si{\kilo\newton\per\meter} &4.7-18 & 9  &  {\cite{Mueller_2016,Smith_2010}}    \\
\hspace{1mm} leg length & $\mathit{l_{0}}$  & \si{\meter}   &  1-1.3 &  1 & {\cite{Abourachid_2000_R, Hutchinson_2004_I,Smith_2010}}  \\
\hspace{1mm} leg angle at TD & $ \mathit{\theta_{L}^{TD}}$  & (\si{\degree}) & 40-68 & $\mathit{f_{A}(\dot{x})}${\protect\footnotemark} &  {\cite{Abourachid_2000_R,Andrada_2014,Gatesy_1999,Mueller_2016}}   \\
\hspace{1mm} dist. Hip-CoM{\protect\footnotemark} & $\mathit{r_{HC}}$  & \si{\meter} & 0.2-0.26 & 0.2 &  {\cite{Abourachid_2000_R,Andrada_2014_A,Fedak_1982,Smith_2010}}    \\ 
\end{tabular}
\end{adjustbox}
\addtocounter{footnote}{-3}
\footnotetext{\,The literature values of the model parameters are presented for an ostrich. The chosen values match an ostrich of \SI{80}{\kilogram}.}
\addtocounter{footnote}{1}
\footnotetext{\,The trunk angle values are calculated from the mean pelvis and hip joint angles from the indicated literature.}
\addtocounter{footnote}{1}
\footnotetext{\,The leg angle is a function of forward speed, see \mcref{fig:Gaits_Damping_th0}b.}
\addtocounter{footnote}{1}
\footnotetext{\,We assume the distance between the hip and \CoM to be equal to the femur length.}
\end{table}                                                                                                                                                                                                                                                                                                                                                                                                                                                                                                                                                                         

\section{TSLIP Model for Avians}\label{sec:Modifications}
The model configuration in \cref{sec:SimModel} is used in \cite{Drama_2019} for the human morphology.  In this section, we underline the modifications made in the \TSLIP model and control, in order to accommodate the avian morphology. First we justify our choice of using a bilinear leg damper with the avian gait data. Second we explain the changes in the control strategy of the leg angle at touch-down and the condition for the leg take-off. Lastly, we clarify the basis for defining \VP position w.r.t. the body and world frames for \VPa and \VPb, respectively.

\subsection{Adaptations in the Model}\label{subsec:ModelMod}
\subsubsection{Bilinear Damping: }\label{subsubsec:BilinearDamping}
\begin{figure}[!tb]
\centering
\begin{annotatedFigure}
{\includegraphics[width=\columnwidth] {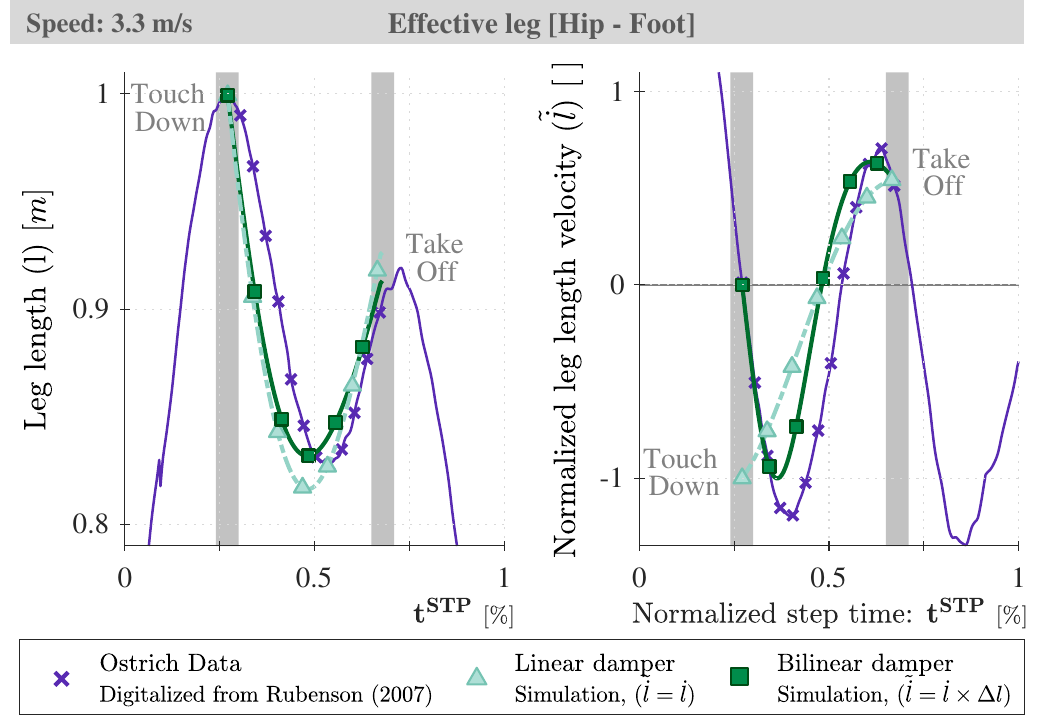}}
	\customSublabelBasic{a)}{0.035,0.86}{color_mygray}
	\customSublabelBasic{b)}{0.49,0.86}{color_mygray}
\end{annotatedFigure}
\caption{The leg length (a) and leg length velocity (b) data of an ostrich and \TSLIP simulation. The \TSLIP model with a linear leg damper ($F_{d}=c\times\dot{l}$) generates non-zero leg length velocity at touch-down event ($\!${\protect \markerDmpLin}$\!$). It results in non-zero damping force at leg touch-down. It is not realistic to have a damping force when the leg is at its rest length. In addition, the simulated length leg velocity does not match the avian gait data({\protect \markerDamperData}). Therefore, we use bilinear damper that accounts for both leg length velocity $\dot{l}$ and its deflection $\Delta l$ in the form of $F_{d}=c\times(\dot{l}\times \Delta l)$. As a result, we obtain zero leg length velocity ({\protect \markerDmpBiLin}), and therefore zero damping force at touch-down. The non-zero leg length velocity at take-off is caused by the early leg take-off, which is also observed in avian data. In the plots, leg lengths are offset to the same touch-down value, velocities are normalized. The ostrich data ({\protect \markerDamperData}) is estimated from the joint angles in Figures 6-10 of \cite{Rubenson_2007}, using inverse kinematics.}
\label{fig:BilinearDamping}
\end{figure}

The need for having a leg damper in the our model is related to the energetic consequence of having pronograde trunk orientation. A pronograde trunk with cranial \CoM requires the hip to supply power continuously, so that the trunk does not collapse into flexion. This positive power needs to be dissipated over a step cycle to maintain a constant energy level. The standard method to achieve this is by adding a parallel leg damper in the \TSLIP model~\cite{Andrada_2014,Sharbafi_2013}. However, the linear leg damper generates non-zero leg length velocity at leg touch-down and take-off events in \mcref{fig:BilinearDamping}b ($\!${\protect \markerDmpLin}$\!$), which is not consistent with the avian gait data ({\protect \markerDamperData}) estimated from \cite{Rubenson_2007}. The inaccurate leg length velocity, causes non-zero damping forces at touch-down and take-off events in the simulation. It is not realistic to have a non-zero damping force at leg touch-down, when the leg is at its rest length. In contrast, non-zero damping force is acceptable at leg take-off due to the early leg take-off observed in avian gaits. The early leg take-off can be seen in \mcref{fig:BilinearDamping}a~({\protect \markerDamperData}), where the leg length at take-off is smaller than at touch-down. Consequently, we use the bilinear leg damper in \cite{Drama_2019} that combines the leg length velocity with leg deflection ({\protect \markerDmpBiLin}) to obtain non-zero damping forces at touch-down. 

\subsubsection{Early Leg Take-off in the Avian Model: }\label{subsubsec:EarlyTO}
%
\setcounter{figure}{3} 
 \begin{figure*}[bh!]
\floatbox[{\capbeside\thisfloatsetup{capbesideposition={right,center},capbesidewidth=0.47\columnwidth}}]{figure}[0.47\FBwidth]
  {\caption{The \TSLIP model is shown at touch-down and take-off events for two cases, where the trunk is perturbed {\color{color_mygray} downward} (a-d) and {\color{color_mygray} upward} (e-h). The \VP has to react to the changes in trunk angle in a stabilizing manner. The {\protect \circledcheckmark} sign indicates that the \VP will counterbalance the trunk perturbation by creating a moment around the \CoM ({\protect\arrowCircCcw}, {\protect\arrowCircCw}) in the opposite direction of the trunk motion. {\protect \circledcross} denotes that it will not. When the trunk is perturbed {\color{color_mygray} downward}, \VP defined in body frame can provide counterbalancing action for both \VPa{\protect \circledcheckmark[color_mygray]} and \VPb{\protect \circledcheckmark[color_mygray]}. However for {\color{color_mygray} upward} trunk perturbation the \VPb{\protect \circledcross[color_mygray]}  in body frame have no means of preventing the trunk from flipping upward (g). The \VPb{\protect \circledcheckmark[color_mygray]}  is defined w.r.t. to the world frame, which flips the relative motion btw.~the \VP and trunk, and works for all cases.}\label{fig:TSLIP_Frames}}
 {\begin{annotatedFigure} 
 {\includegraphics[width=1.05\columnwidth]{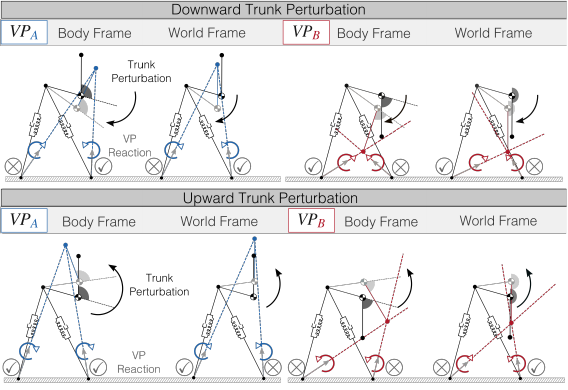}}
 \customSublabelBasic{a)}{0.02,0.85}{color_mygray}
\customSublabelBasic{b)}{0.31,0.85}{color_mygray}
\customSublabelBasic{c)}{0.52,0.85}{color_mygray}
\customSublabelBasic{d)}{0.77,0.85}{color_mygray}
\customSublabelBasic{e)}{0.02,0.365}{color_mygray}
\customSublabelBasic{f)}{0.31,0.365}{color_mygray}
\customSublabelBasic{g)}{0.52,0.365}{color_mygray}
\customSublabelBasic{h)}{0.77,0.365}{color_mygray}
 \end{annotatedFigure}}
\end{figure*}

The standard \TSLIP model terminates the stance phase when the leg reaches to its rest length. Condition 1 causes an unexpected phenomenon for the pronograde TSLIP model, in particular at slow speeds, which is demonstrated in \mcref{fig:TOconditions} for \SI{4}{\metre\per\second}. The \CoM reaches its apex height towards the end of the stance phase, while the leg length is still smaller that its rest length $l_{0}$. Consequently, the stance phase continues and the \CoM starts to accelerate downward until the leg is fully extended (\multirefrange{fig:TOconditions}{c}{d},{\protect \markerLineGrey \protect \markerCrossGreen}\,). In other words, the \CoM behaves like an inverted pendulum (IP) towards end of the stance phase. The flight phase from leg take-off to apex diminishes, and the phase from apex to leg-touch down starts at a lower height with negative vertical velocity. The IP-like behavior is uncharacteristic for the running gaits with spring-mass dynamics. In order to prevent the undesired negative vertical \CoM velocity and negative vertical \GRF after the midstance (MS), we add Cond.~2-3 to the stance phase termination conditions. As a result, the leg length at take-off is shorter than the rest length (\multirefrange{fig:TOconditions}{e}{f},{\protect \markerLineGrey}) and the GRF profiles end suddenly (\multirefrange{fig:TOconditions}{g}{h},{\protect \markerLineGrey}).

\setcounter{figure}{2} 
\begin{figure}[!t]
\centering
\begin{annotatedFigure}
{\includegraphics[width=\columnwidth]{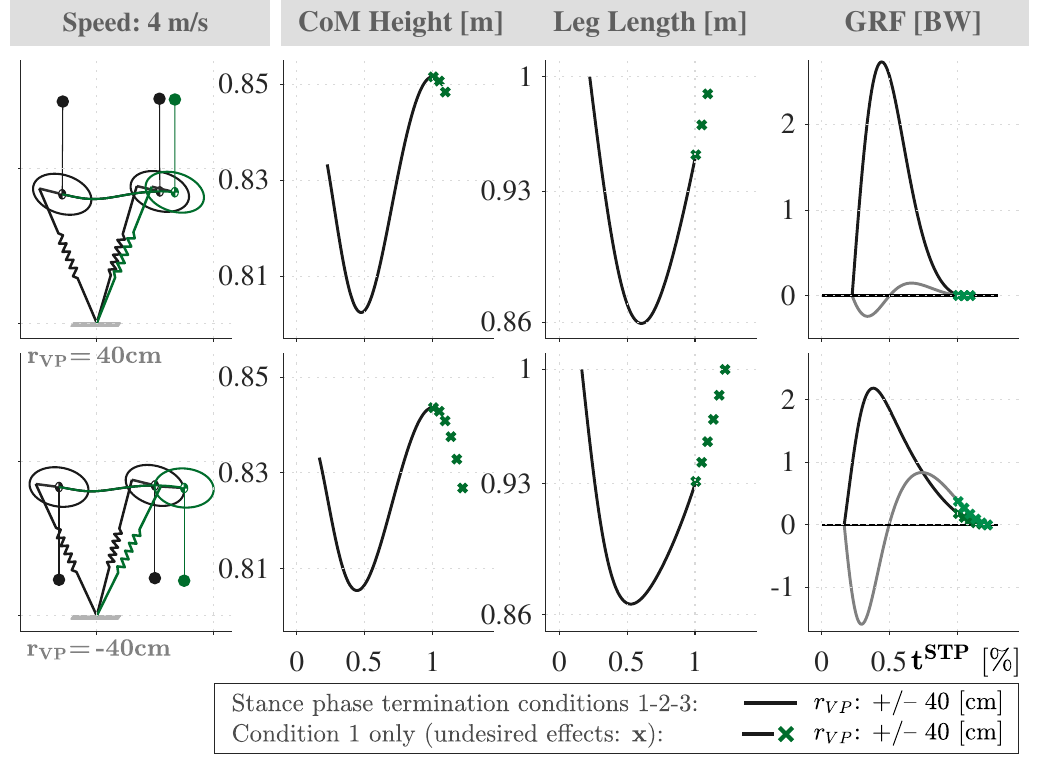}}
\customSublabelBasic{a)}{0.045,0.9}{color_mygray}
\customSublabelBasic{c)}{0.3,0.9}{color_mygray}
\customSublabelBasic{e)}{0.545,0.9}{color_mygray}
\customSublabelBasic{g)}{0.8,0.9}{color_mygray}
\customSublabelBasic{b)}{0.045,0.48}{color_mygray}
\customSublabelBasic{d)}{0.3,0.48}{color_mygray}
\customSublabelBasic{f)}{0.545,0.48}{color_mygray}
\customSublabelBasic{h)}{0.8,0.48}{color_mygray}
\end{annotatedFigure}
\caption{(a-b) Touch-down and take-off events of the \TSLIP model are shown for \VP radii of $\pm$\SI{40}{\centi\meter}. The gait kinematics (c-f) and kinetics (g-h) are drawn for two case scenarios, where the stance phase terminates at conditions~1-2-3 ({\protect \markerLineGrey}) and condition 1 only ({\protect \markerLineGrey \protect \markerCrossGreen}). In general, the leg take-off condition~1 ($l \mygeq l_{0}$) is used in the TSLIP model to terminate the stance phase. However, condition~1 alone leads to gaits, where the \CoM to reaches its apex and gains negative acceleration towards the end of the stance phase. This IP-like behavior is marked with crosses ({\protect \markerCrossGreen}) in (c-d) and is not a part of the running gait with spring-mass dynamics. Consequently, we extend the leg take-off event condition to 2-3 to prevent negative vertical speed of the \CoM after the midstance. The extended conditions result in early leg take-off (e-f) and a cut-off in ground reaction forces (g-h).} 
\label{fig:TOconditions}
\end{figure}
 
In summary, the stance phase is terminated when one of the three conditions below holds. Consequently, the leg takes off earlier, before the \CoM reaches to its apex, as shown in \multirefrange{fig:TOconditions}{c}{d} ({\protect \markerLineGrey}).
\begin{compactitem}[]
\setlength{\itemsep}{0.1cm}%
\item \begin{event}\label{condition:1_TO} The leg reaches its rest length:~$l \mygeq l_{0}$.
\end{event}
\item \begin{event}\label{condition:2_TO}
The vertical \CoM speed reaches zero after midstance. I.e., $\dot{y}_{C} \myleq 0$.
\end{event}
\item \begin{event}\label{condition:3_TO}
The vertical \GRF reaches zero. I.e., $GRF_{vert.} \myleq 0$, which presents the unilateral constraint.
\end{event}
\end{compactitem}

\subsection{Adaptations in the Control Strategy}\label{subsec:ControlMod}
We use a linear controller to regulate the leg angle at touch-down, which is a function of the forward speed and apex height \cite{Raibert_1986,Drama_2019}. In the avian \TSLIP model, we add a linear dependence of the body angle at apex on the leg angle at touch-down to bound the magnitude of the trunk oscillations. We explain the selection of the controller gains in \cref{subsec:ExpSetup}.

We determine the hip torque using a \VP concept, where the \VP creates a passive control mechanism that guides the \GRF vectors and counteracts the trunk pitch motion (see {\protect \circledcheckmark} in \mcref{fig:TSLIP_Frames}). If the \VP does not provide countering motion {\protect \circledcross}, the trunk would either flip back or collapse into flexion, and the motion would fail. A major factor that determines this reaction is the coordinate system in which \VP is defined. The \VP frame should be selected so that the resultant \GRF creates a moment around the \CoM in the opposite direction of the trunk motion during at least some part of the stance phase. 

In the standard \TSLIP model, the \VP is defined w.r.t. the body frame. When the trunk is perturbed {\color{color_mygray} downward} as in \multirefrange{fig:TSLIP_Frames}{a}{d}, the \VP can generate instances with counteracting moment around the \CoM for both \VPa{\protect \circledcheckmark[color_mygray]} and \VPb{\protect \circledcheckmark[color_mygray]}. When the trunk is perturbed {\color{color_mygray} upward}, it is possible for \VPa  {\protect \circledcheckmark[color_mygray]} (\multirefrange{fig:TSLIP_Frames}{e}{f}), but \emph{not} feasible for \VPb{\protect \circledcross[color_mygray]} (\multirefcomma{fig:TSLIP_Frames}{g}{h}), to counteract the upward moving trunk during stance phase (i.e.,~{\protect \circledcross[color_mygray] in }\mcref{fig:TSLIP_Frames}g). For \VPb, the set \VP location cannot provide postural equilibrium. Since the \VP angle is non-adaptive, there is no means of recovering from a upward trunk motion. We propose defining the \VPb w.r.t. the world frame, which flips the direction of the \VP location change w.r.t. the trunk motion. This way the trunk can stabilize {\color{color_mygray} upward} perturbations of trunk and can obtain steady state solutions for \VPb. 

\subsection{Gait Generation}\label{subsec:ExpSetup}

In our simulation setup, we sweep \VP targets over $r_{VP} \myeq \pm [0,\, 20,\, 40,\, 60]$ \si{\centi\meter}{\protect\footnote{\,The parameter sweep for \VPb at \SI{10}{\metre\per\second} ends at $r_{VP} \myeq$\SI{52}{\centi\metre} due to instability caused by high angular trunk acceleration.}}{\protect\footnote{\,The negative $r_{VP}$ denotes that the \VP is below the~\CoM.}} to create trunk oscillations with magnitude up to \SI{10}{\degree} (see \mcref{fig:VP}), which is the maximum value reported for avian running in literature \cite{Abourachid_2011,Gatesy_1999, Jindrich_2007, Rubenson_2007}. We define the \VP angle $\theta_{VP}$ in body frame for \VPa and in world frame for \VPb, due to reasons given in \cref{subsec:ControlMod}. The \VP concept requires the points to reside on the vertical axis passing through the \CoM for steady state gaits~\cite{Sharbafi_2013}. Therefore, \VP angles are set to  $\theta_{VP} \myeq \minus \theta_{C}$ and $\theta_{VP} \myeq \minus \pi$ respectively. In addition, we set the desired mean body pitch angle to \SI{100}{\degree}, which is in accordance with the reported values in literature (see \cref{tab:ModelPrm}).

\setcounter{figure}{4} 
\begin{figure}[!tb]
\centering
\begin{annotatedFigure}
 {\includegraphics [width=\columnwidth] {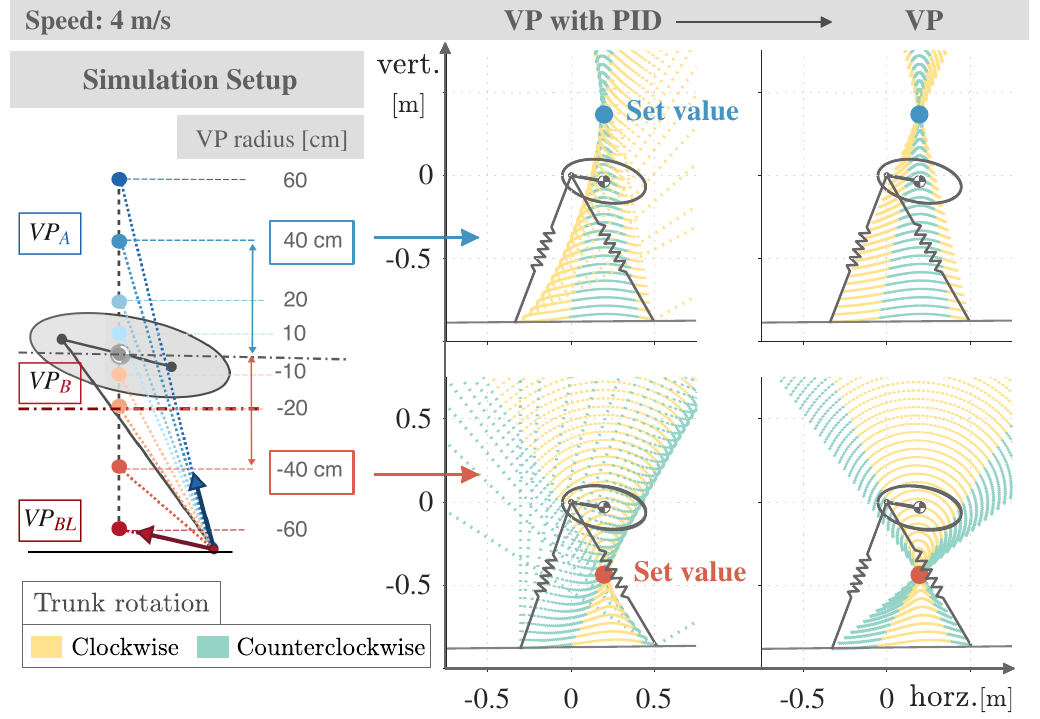}}
  \customSublabelBasic{a)}{0.04,0.81}{color_mygray}
  \customSublabelBasic{b)}{0.46,0.915}{color_mygray}
   \customSublabelBasic{d)}{0.71,0.915}{color_mygray}
  \customSublabelBasic{c)}{0.46,0.435}{color_mygray}
   \customSublabelBasic{e)}{0.71,0.435}{color_mygray}
\end{annotatedFigure}
\caption{The simulation setup (a) and resulting \GRF patterns (b-e) for the avian \TSLIP model. It is difficult to find feasible gaits for the \VP scheme, as the control is sensitive to the changes in the initial state and model parameters. To guide the gait search, we use the iterative control framework presented in \cite{Drama_2019} (refer to \cref{sec:SimModel}). The framework has an initial step, which combines the \VP controller with a PID controller on trunk pitch angle. The resultant gaits do not have a focused \GRF profile (b-c). Then, we reduce the PID gains gradually, until the \GRF intersect at a single point (d-e).}
\label{fig:VP}
\end{figure}

\begin{figure}[!tb]
\centering
\begin{annotatedFigure}
 {\includegraphics[width=1\columnwidth]{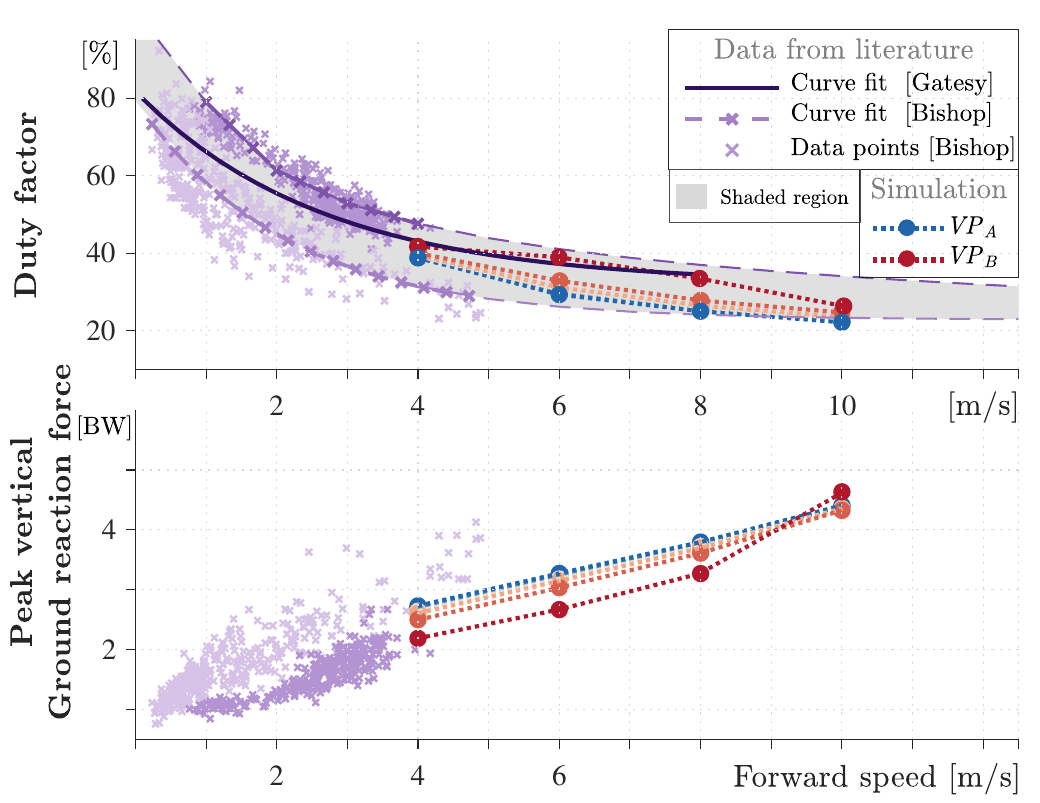}}
    \customSublabelBasic{a)}{0.02,0.96}{color_mygray}
     \customSublabelBasic{b)}{0.02,0.5}{color_mygray}
\end{annotatedFigure}
\caption{Control gains and damping coefficients of the \TSLIP model are tuned, such that the resultant gaits yield duty factor (a) and peak vertical \GRF (b) values similar to ones reported in the biomechanical measurements in literature. Data points for the duty factor are retrieved from \cite{Bishop_2018}. They cover the speeds \SIrange[range-phrase=-,range-units=single]{0.25}{5}{\meter\per\second}, and show two trends: a lower valued curve for small birds ({\protect \markerGfowl}) and a higher valued curve for ratites ({\protect \markerGfowlR}). We fit two curves to these data sets, which determine the lower and upper bound of the grey shaded region.  We use the duty factor curve digitalized from \cite{Gatesy_1991} to extend this region to higher speeds up to \SI{8}{\meter\per\second} ({\protect \markerLit}). The duty factor of our simulated gaits ({\protect \markerVPa},{\protect \markerVPb}) lies in the grey region and decreases with the forward speed. We monitor the peak vertical \GRF values in a similar fashion.}
\label{fig:Lit_duty}
\end{figure}

We select the model parameters and controller gains of our avian model so that the resulting gaits follow the trends observed in biomechanics. We choose to match the duty factor{\protect\footnote{\, Duty factor is the ratio of the leg contact time to the stride~period}}, the \GRF profile and the leg angle at touch-down; which are a function of the forward speed. As forward speed increases, the duty factor gets smaller, magnitude of the \GRF gets higher, and leg touch-down angle gets smaller  \cite{Abourachid_2000_R,Bishop_2018,Blum_2011,Gatesy_1991,Mueller_2016,Rubenson_2004}.
We tune the damping coefficient and control gains of our model so that the resulting gait behavior is feasible and follows the trend reported for avian locomotion, {\protect \markerGfowl} in \mcref{fig:Lit_duty}. Simulated gaits ({\protect \markerVPa},{\protect \markerVPb}) have a duty factor of \SIrange[range-phrase=-,range-units=single]{40}{60}{\percent} in \mcref{fig:Lit_duty}a, and peak vertical {\GRF} of \SIrange[range-phrase=-,range-units=single]{2}{5}{BW} in \mcref{fig:Lit_duty}b. The leg touch-down angle and damping coefficient are in the range of \SIrange[range-phrase=-,range-units=single]{60}{40}{\degree} and \SIrange[range-phrase=-,range-units=single]{6}{1.5}{\kilo\newton\second\per\meter} and decrease with speed (see \mcref{fig:Gaits_Damping_th0}a). 

\section{Simulation Results}\label{sec:SimResults}
In this section, we analyze the results of our simulation setup to investigate the effect of trunk oscillations.

\subsection{Asymmetries in the Kinetics and Kinematics}\label{subsec:Asymmetries}
The trunk inclination ($\theta_{C}$ in \mcref{fig:TSLIP}b) introduces asymmetries in the system, which are reflected in the leg dynamics and \GRF profiles \cite{Andrada_2014, Blickhan_2015}. This effect is pronounced for the avian model, where the \CoM is placed cranially with an inclination of $100^\circ$.  

\begin{figure}[!tb]
\centering
\begin{annotatedFigure}
 { \includegraphics [width=1\columnwidth] {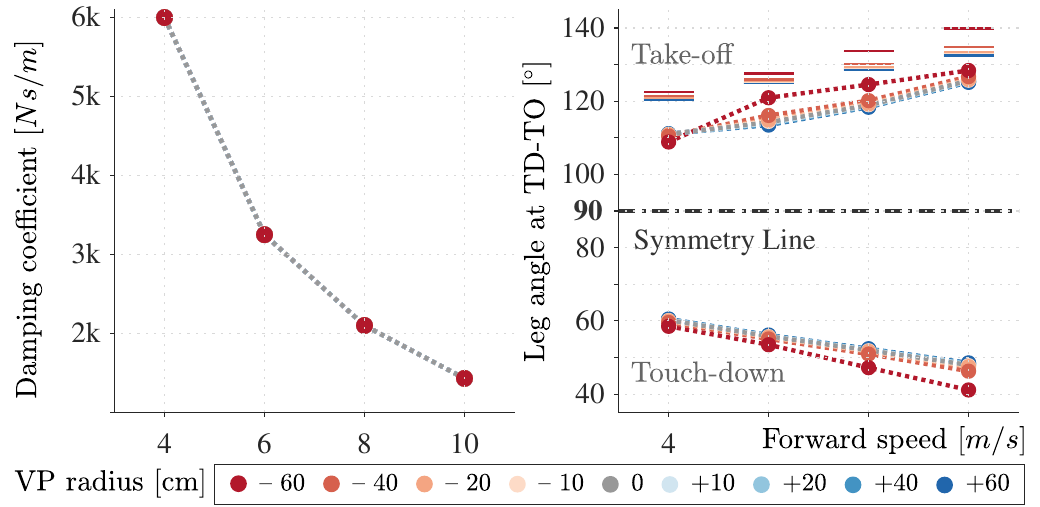}}
      \customSublabelBasic{a)}{0.03,0.975}{color_mygray}
     \customSublabelBasic{b)}{0.515,0.975}{color_mygray}
\end{annotatedFigure}
\caption{The damping coefficient (a) and leg touch-down angle (b) of the simulated gaits are inversely proportional to the forward running speed. Leg touch-down and take-off angles are not symmetric about the vertical axis (i.e.,~symmetry line, $90^{\circ}$). The solid lines show what the leg take-off angles would be, if the gaits were to be symmetrical. The leg angle asymmetry is a consequence of the pronograde trunk orientation in our model.}
\label{fig:Gaits_Damping_th0}
\end{figure}

\begin{figure}[!tb]
\centering
\begin{annotatedFigure}
  {\includegraphics [width=1\columnwidth]{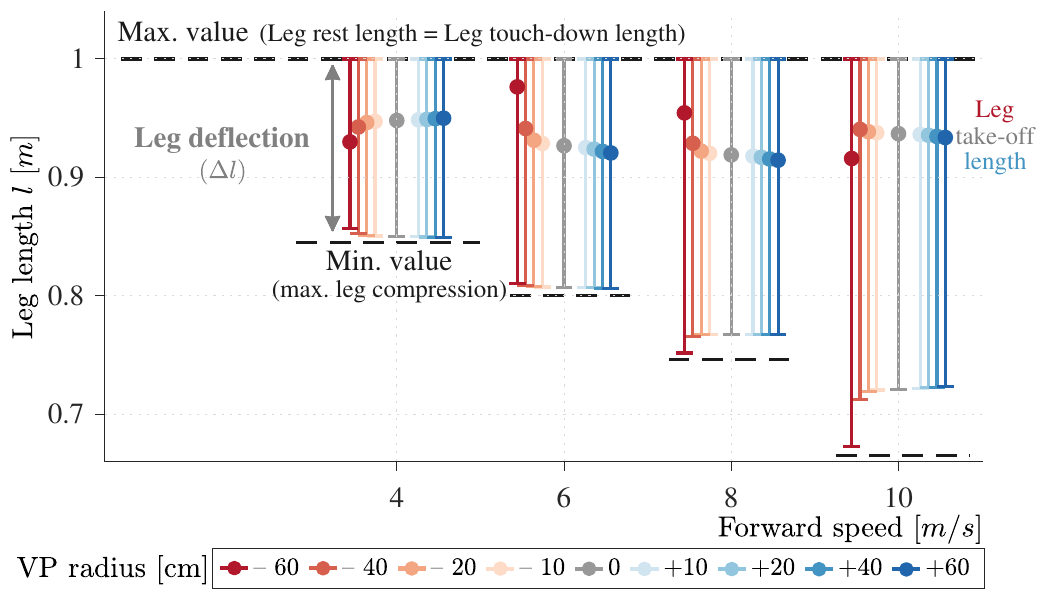}}
\end{annotatedFigure}
\caption{The min.~and max.~values of the leg length are plotted as bar plot, where the leg take-off length is marked with dots ({\protect \markerVPa},{\protect \markerVPb}). We observe an early leg take-off with a shorter leg, which is approximately \SIrange[range-phrase=-,range-units=single]{4}{8}{\percent} of the leg rest length $l_{0}$. The leg deflection corresponds to the length of each bar and increases from \SIrange[range-units=single,range-phrase=\ to\ ]{15}{30}{\percent} of $l_{0}$ with the running speed.}
\label{fig:Gaits_LegLength_Asym}
\end{figure}

The asymmetry is apparent for the leg angles at touch-down and take-off in \mcref{fig:Gaits_Damping_th0}b, where the leg takes off before the leg length reaches to its resting value, $l_{0}$. The difference between leg lengths at touch-down and take-off varies between \SIrange[range-phrase=-,range-units=single]{4}{8}{\percent} of the $l_{0}$ (see max. length and ({\protect \markerVPa,\protect \markerVPb }) in \mcref{fig:Gaits_LegLength_Asym}). The stance phase is terminated early either due to loss of foot contact caused by the vertical \GRF decaying to zero, or due to the \CoM reaching to its apex height (\cref{condition:2_TO}-\labelcref{condition:3_TO}). As a result of the early leg take-off, the leg spring is unable to inject all stored energy. Effectively, the leg spring removes energy from the system, details of which we discuss in \cref{subsubsec:WorkLegHip}.

\begin{figure}[tb!]
\centering
\begin{annotatedFigure}
 { \includegraphics [width=1\columnwidth]{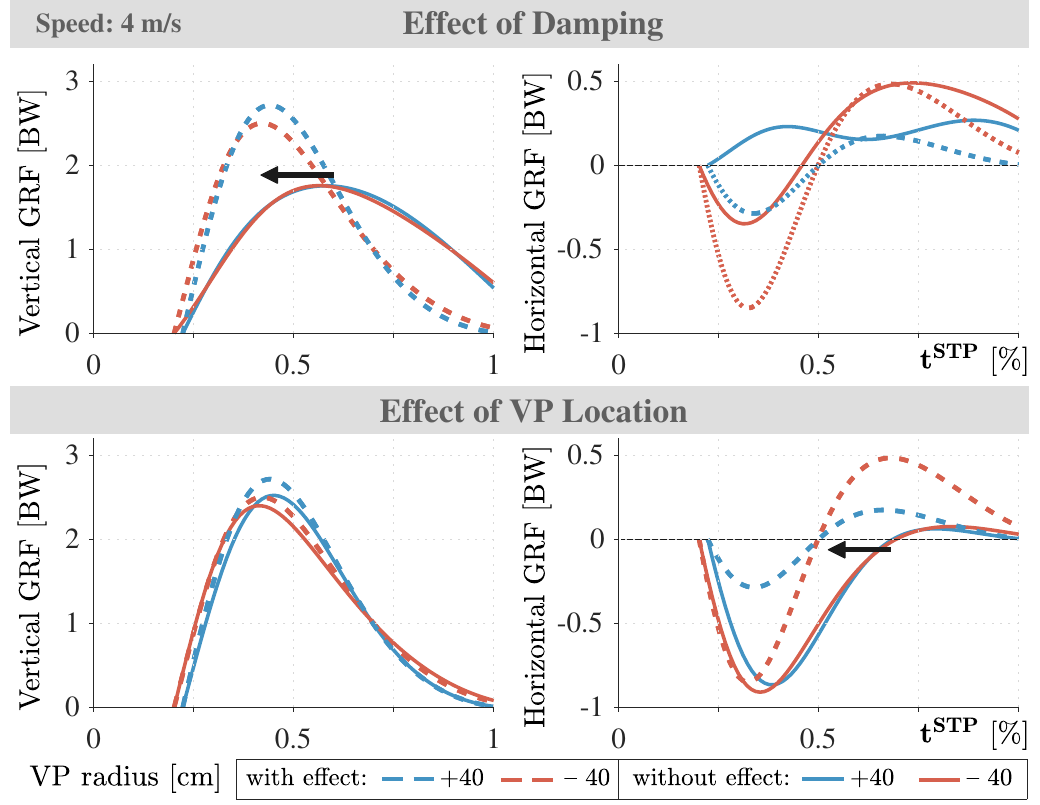}}
      \customSublabelBasic{a)}{0.13,0.9}{color_mygray}
     \customSublabelBasic{b)}{0.63,0.9}{color_mygray}
     \customSublabelBasic{c)}{0.13,0.4}{color_mygray}
      \customSublabelBasic{d)}{0.63,0.4}{color_mygray}
\end{annotatedFigure}
\caption{The dotted lines are the \GRF profiles for the gaits having a \SI{40}{\centi\meter} \VP target and \SI{4}{\meter\per\second} speed. The \GRF is the sum of the axial force created by the leg ($\prescript{}{F}{\mathbf{F}}_{a}$)  and the tangential force created by the hip ($\prescript{}{F}{\mathbf{F}}_{t}$). In (a-b, solid lines), we subtract the force generated by the leg damper ($\mathbf{F}_{dp} \myeq \prescript{}{F}{\mathbf{F}}_{a}-\mathbf{F}_{sp}$) form the \GRF. We observe that the damping causes a left skew in {vertical \GRF} (a,{\protect \arrowPink}), and has no significant effect for the horizontal \GRF. In (c-d), we subtract the tangential force generated by the hip  ($\prescript{}{F}{\mathbf{F}}_{t}$) from the \GRF. We observe that the hip torque causes a left skew towards left in horizontal \GRF (d,{\protect \arrowPink}), and has no significant effect on the vertical \GRF. }
\label{fig:GRF_Asym}
\end{figure}

In our gait framework, we obtain left-skewed \GRF profiles shown in \mcref{fig:GRF_Asym}, which is consistent with the avian locomotor data in \cite{Andrada_2014, Bishop_2017, Clemente_2018}. In the following, we investigate potential reasons. First, we subtract the component of the \GRF created by the leg damper ($\mathbf{F}_{dp} \myeq \prescript{}{F}{\mathbf{F}}_{a} \minus \mathbf{F}_{sp}$) from the total \GRF in \multiref{fig:GRF_Asym}{a}{b} (solid lines). We observe that the damping force skews vertical \GRF to the left (\mcref{fig:GRF_Asym}a, dotted lines,{\,\protect \arrowPink}). The effect of the damping is visible mainly in vertical \GRF profile, because the leg force makes up most of the vertical \GRF and only a minor part of horizontal \GRF. 

Next, we subtract the component of the \GRF produced by the hip torque ($\prescript{}{F}{\mathbf{F}}_{t}$) in \multiref{fig:GRF_Asym}{c}{d} (solid lines). The hip torque contributes mainly to the horizontal \GRF and shifts the zero crossing of the horizontal \GRF profile to the left (\mcref{fig:GRF_Asym}d, dotted lines,{\,\protect \arrowPink}). The integral of the horizontal \GRF curve (the area) is the the fore-aft impulse, which corresponds to the forward acc/deceleration of the main body. 
The horizontal \GRF created by the leg force causes larger negative horizontal impulse than the positive one, which would effectively decelerate the main body. An example is presented in \mcref{fig:GRF_Asym}d, where the positive and negative impulses generated by the horizontal \GRF (solid lines) amount to \SIrange[range-phrase=/,range-units=single]{3}{\minus 48}{\newton\second} for \VPa~(blue) and \SIrange[range-phrase=/,range-units=single]{4}{\minus 52}{\newton\second} for \VPb~(red) respectively.
The hip torque (dashed line) produces forces to ensure sufficient forward acceleration to generate the motion. In other words, it creates equal positive and negative fore-aft impulses by shifting the zero crossing of horizontal \GRF to the left.

\begin{figure}[!tb]
\centering
\begin{annotatedFigure}
  {\includegraphics [width=1\columnwidth]{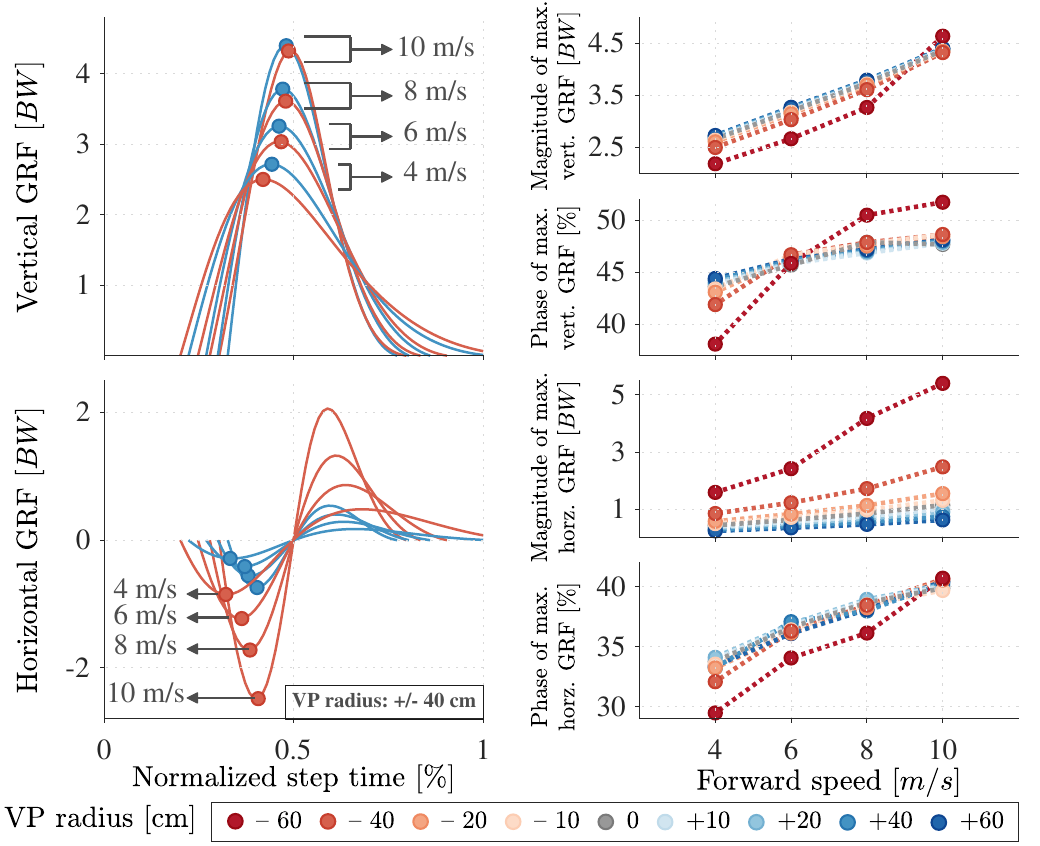}}
        \customSublabelBasic{a)}{0.14,0.9}{color_mygray}
     \customSublabelBasic{c)}{0.48,0.95}{color_mygray}
     \customSublabelBasic{d)}{0.48,0.7}{color_mygray}
      \customSublabelBasic{b)}{0.14,0.48}{color_mygray}
      \customSublabelBasic{e)}{0.48,0.5}{color_mygray}
      \customSublabelBasic{f)}{0.48,0.25}{color_mygray}
\end{annotatedFigure}
\caption{The \GRF patterns  are shown  in (a-b) for \SI{40}{\centi\meter} \VP target at speeds \SIrange[range-phrase=-,range-units=single]{4}{10}{\meter\per\second}. A~\VPa~({\protect \markerVPa}) yields higher peak vertical \GRF and lower peak horizontal \GRF, compared to \VPb~({\protect \markerVPb}). The~magnitude of the peak \GRF increases with speed, which is quantified in (c,e). The phase of the peak \GRF (d,f)  is calculated w.r.t. the step time and it increases with speed. In other words, the \GRF become more symmetrical at higher speeds.} 
\label{fig:GRF_dx}
\end{figure}

We then investigate how the skew of the \GRF profile changes with the forward speed in \mcref{fig:GRF_dx}. As the forward speed gets higher, the magnitude of peak horizontal and vertical \GRF increase between \SIrange[range-phrase=-,range-units=single]{2.5}{4.5}{BW} and \SIrange[range-phrase=-, ,range-units=single]{0.5}{2}{BW}, respectively (\multirefcomma{fig:GRF_dx}{c}{e}). The \emph{phase} of peak \GRF is calculated with respect to the gait cycle (marked with ({\protect \markerVPa},{\protect \markerVPb}) in \multirefrange{fig:GRF_dx}{a}{b}). The phase of peak horizontal and vertical \GRF increases between \SIrange[range-phrase=- ,range-units=single]{42}{84}{\percent} and \SIrange[range-phrase=-,range-units=single]{32}{40}{\percent}, respectively (\multirefcomma{fig:GRF_dx}{d}{f}). The increase in phase indicates that gaits become more symmetric at high speeds. 

In terms of the \VP location, \VPa and \VPb have similar phase values for peak {\GRF}. \VPb yields lower peak vertical \GRF and higher peak horizontal \GRF magnitudes shown in  \multirefrange{fig:GRF_dx}{a}{b}, which are associated with high duty factor and high horizontal accelerations, respectively.  In other words, \VPb makes the \CoM brake and accelerate more during stance phase in the horizontal direction. 

Apart from the asymmetries mentioned, we detect no take-off--apex phase occurring at low speeds, around \SI{4}{\meter\per\second} with high duty factor (see grey shaded area in \multirefrange{fig:thB}{a}{b}, or \multirefrange{fig:TOconditions}{c}{d}). The \CoM reaches its peak height at the end of the stance phase and the next step begins immediately after, which is enabled with the take-off condition \cref{condition:2_TO}-\labelcref{condition:3_TO} in \cref{subsec:ModelMod}.

\subsection{Trunk Pitch Oscillations}\label{subsec:TrunkPitchOscillations}
\begin{figure}[!tb]
\centering
\begin{annotatedFigure}
  {\includegraphics [width=1\columnwidth]{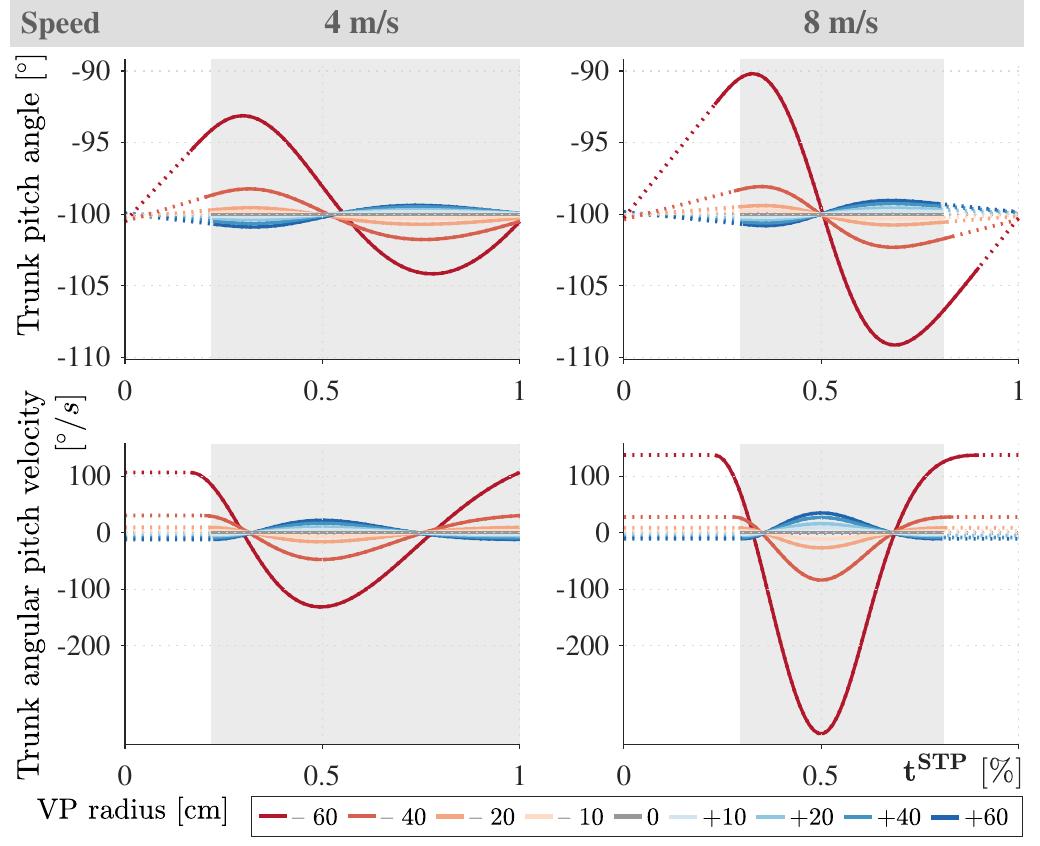}}
        \customSublabelBasic{a)}{0.15,0.91}{color_mygray}
     \customSublabelBasic{b)}{0.63,0.91}{color_mygray}
     \customSublabelBasic{c)}{0.15,0.51}{color_mygray}
      \customSublabelBasic{d)}{0.63,0.51}{color_mygray}
\end{annotatedFigure}
\caption{The trunk pitch angle (a-b), and trunk pitch velocity (c-d) are shown for \SIrange[range-units=single]{4}{8}{\meter\per\second}. \VPa leads to upward trunk motion during the stance phase of running, whereas \VPb produces a reverse movement. The trunk angular excursion (i.e.,~the absolute value of the difference between min.~and max.~trunk angles) and the peak angular pitch velocity increases with the \VP radius. The stance phase is shaded in gray.}
\label{fig:thB}
\end{figure}

In our simulations, \VPb leads to {\protect \markerSineB}-shaped, downward trunk pitch motion in during the stance phase, which are similar to the avian gait characteristics reported in \cite{Hancock_2014, Jindrich_2007}. \VPa yields {\protect \markerSineA}-shaped, upward trunk motions in \multiref{fig:thB}{a}{b}. For equal \VP radius, \VPb causes larger pitch oscillations. The magnitude of the oscillations and the angular velocity increase with the
 \VP radius for both \VPa and \VPb (\multirefrange{fig:thB}{c}{d}). 

\begin{figure}[!bt]
\centering
\begin{annotatedFigure}
 { \includegraphics [width=\columnwidth]{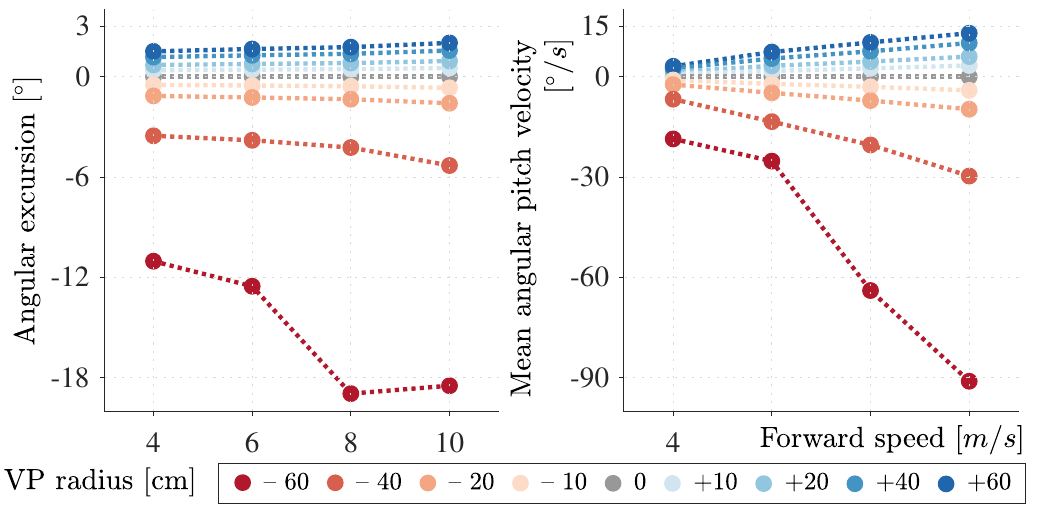}}
        \customSublabelBasic{a)}{0.025,0.975}{color_mygray}
     \customSublabelBasic{b)}{0.515,0.975}{color_mygray}
\end{annotatedFigure}
\caption{The trunk angular excursion (i.e.,~the absolute value of the difference between min.~and max.~trunk angles) and mean angular pitch velocity increase with forward speed and increasing \VP radius.}
\label{fig:thBdthB_dx}
\end{figure}

When the running speed increases from \SIrange[range-units=single]{4}{10}{\meter\per\second}, the trunk angular excursion increases up to \SI{2}{\degree} for \VPa and \SI{18}{\degree} for \VPb (see \mcref{fig:thBdthB_dx}a). The mean trunk angular velocity increases up to \SI{13}{\degree\per\second} for \VPa and $\minus$\SI{91}{\degree\per\second} for \VPb (see \mcref{fig:thBdthB_dx}b). 

\subsection{Energy Considerations} \label{subsec:Energy}
In this section, we investigate how the leg and hip contributes to the system's energy balance, and how the \CoM energy evolves over step time. Moreover, we provide the mechanical cost of transport to allow comparison to the literature.
%
To clarify, we use the term {\color{color_VPcom} \emph{$positive/negative$ work}} to address \emph{the amount} of energy created/absorbed by the leg force or the hip torque. We define the {\color{color_VPcom} \emph{energy fluctuation}} ${\mathcolor{color_VPcom}{ \left( \Delta \right) }}$ as the difference between maximum and minimum values of any energy type. 
%

\subsubsection{Work Distribution Between Leg and Hip}\label{subsubsec:WorkLegHip} \hfill 

\begin{figure}[!tb]
\centering
\begin{annotatedFigure}
  {\includegraphics [width=\columnwidth] {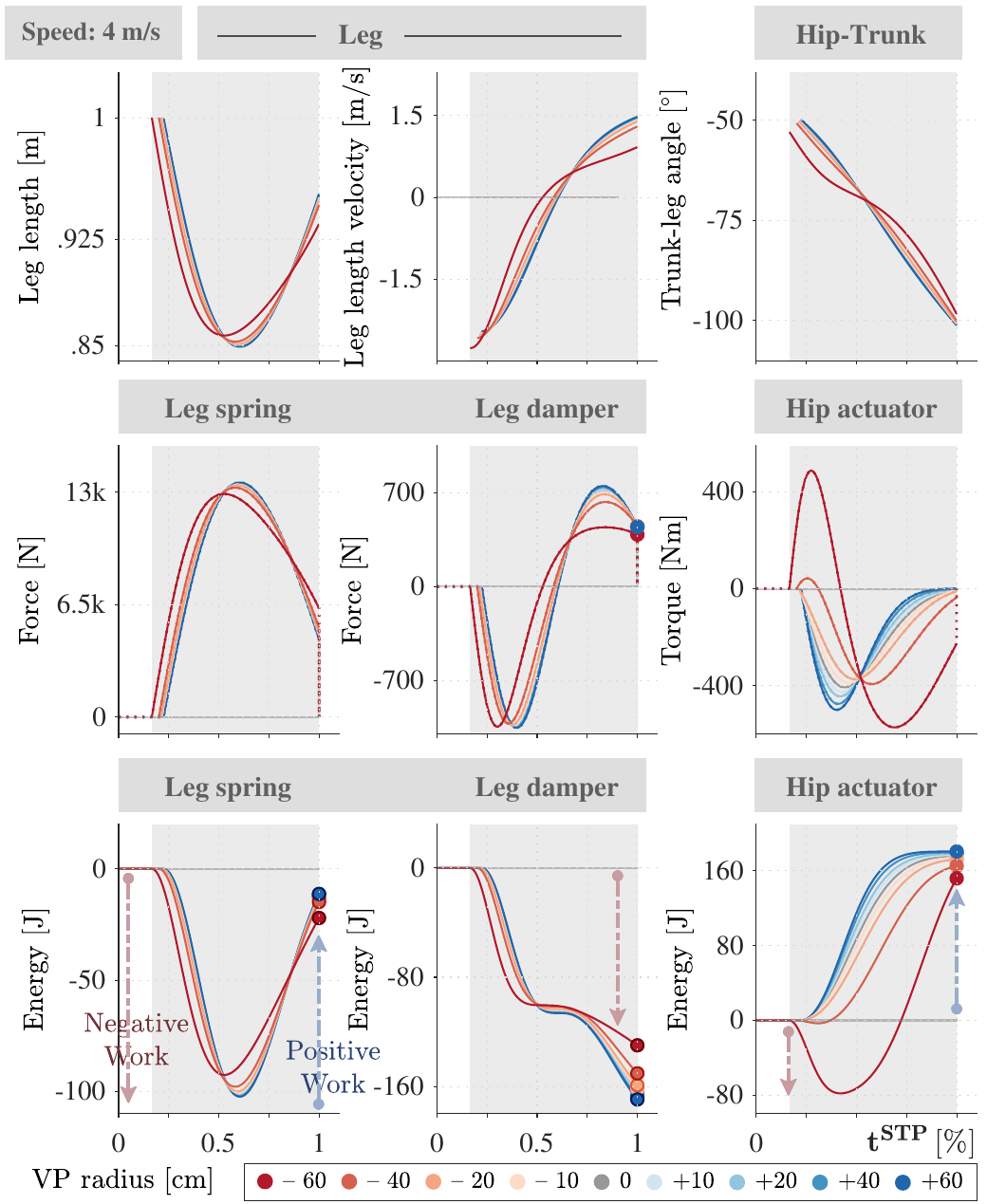}}
    \customSublabelBasic{a)}{0.1,0.935}{color_mygray}
\customSublabelBasic{b)}{0.4,0.935}{color_mygray}
  \customSublabelBasic{c)}{0.73,0.935}{color_mygray}
\customSublabelBasic{d)}{0.1,0.62}{color_mygray}
  \customSublabelBasic{e)}{0.4,0.62}{color_mygray}
\customSublabelBasic{f)}{0.73,0.62}{color_mygray}
  \customSublabelBasic{g)}{0.1,0.305}{color_mygray}
\customSublabelBasic{h)}{0.4,0.305}{color_mygray}
  \customSublabelBasic{i)}{0.73,0.305}{color_mygray}
  \end{annotatedFigure}
\caption{The leg spring force (d), damper force (e), hip actuator torque (f) and their respective energies (g-i) for \VPa and \VPb configurations with varying \VP radius. The hip needs to perform net positive work (i.e.,~inject energy) to hold the pronograde trunk in position and assist the \CoM motion (i,{\,\protect\arrowWorkP\,}\,{\,\protect\arrowWorkN\,}). The net work performed on the system has to remain zero to obtain steady state motion. The task of energy depletion is achieved primarily by the leg damper (h, {\,\protect\arrowWorkP\,}), and partly by the leg spring (g,{\,\protect\arrowWorkP\,}\,{\protect\arrowWorkN\,}) owing to the early take-off (a). \VPa causes higher positive net hip work and negative leg damper work (i,h,{\protect \markerVPa}), compared to \VPb~({\protect \markerVPb}). In contrast, \VPa yields lower negative net spring work (g,{\protect \markerVPa}).}
\label{fig:Kinetics}
\end{figure}

\paragraph{Temporal Analysis: }\label{subsubsecpara:WorkLegHip_t} 
We explore how the leg force, hip torque, and their respective energies evolve over the course of the stance phase in \mcref{fig:Kinetics} and \multirefrangewo{fig:Energy_dx}{j}{l}. The leg spring deflects and stores energy during the first half of the stance phase (\mcref{fig:Kinetics}g, \labelcref{fig:Energy_dx}j,~{\protect\arrowWorkN\,}). It extends and returns this energy back to the main body in the second half (\mcref{fig:Kinetics}g,~{\protect\arrowWorkP\,}). However, owing to the early leg take-off, the spring is not able to recoil completely at the end of the stance phase (\mcref{fig:Kinetics}a) and return all the energy it absorbed (refer to \cref{subsec:Asymmetries} and see  \mcref{fig:Kinetics}g). Consequently, the spring has a net effect of removing energy from the system, which is indicated by the arrows (\mcref{fig:Kinetics}g,~{\,\protect\arrowWorkP\,}\,{\protect\arrowWorkN\,}). 
Concerning the leg damper, early leg take-off interrupts the energy absorption of damper (\mcref{fig:Kinetics}h,\labelcref{fig:Energy_dx}k,~{\protect\arrowWorkN\,}) and makes the damping force end abruptly at a non-zero value (\mcref{fig:Kinetics}e,~{\!\protect\markerVPa\protect\markerVPb}). 

The hip actuator has two purposes: to compensate for the energy losses of the leg and to provide positive net work to support the forward leaning trunk against gravity. The posterior placement of the hip w.r.t.~the \CoM for a pronograde trunk necessitates high positive hip work to hold the heavy trunk \cite{Blickhan_2015,Andrada_2014_A}.
Accordingly, we see in \mcref{fig:Kinetics}f that the hip torque is always negative for \VPa, and \VPb with radii smaller than \SI{30}{\centi\metre}. The hip solely injects energy to the system (\mcref{fig:Kinetics}i,\labelcref{fig:Energy_dx}l,~{\protect\arrowWorkP\,}).   
If the \VPb radius is larger than \SI{30}{\centi\metre} (\VPbl), the \VP is set below the leg axis at touch-down. In this case, the hip starts producing positive torque in the first half of the stance phase (\mcref{fig:Kinetics}f), where the hip depletes energy (\mcref{fig:Kinetics}i,~{\protect\arrowWorkN\,}) and assists the leg force to decelerate the body. Beyond this initial negative work, the hip still has to produce net positive work to offset the energy absorbed by the leg (\mcref{fig:Kinetics}i,~{\protect\arrowWorkP\,}{\,\protect\markerVPb}). 
%

The requirement for the net positive hip work comes from the \emph{steady state condition}, where the net work done on the system must be zero{\protect\footnote{\,Asymmetric or period-2 gaits are not considered.}. Otherwise the body would accelerate or decelerate. Following this, we see that the energy injected by the hip actuator is equal to the energy absorbed by the leg in \mcref{fig:WorkDistribution}. The leg produces positive work via the spring and negative work through both the spring and damper, which results in a net negative work (\multirefrange{fig:WorkDistribution}{a}{b}). The contribution of the leg spring to the overall energy removal is relatively small and amounts to less than \SI{5}{\percent} (\multirefrange{fig:Kinetics}{g}{i}).

\begin{figure}[!tb]
\centering
  \begin{annotatedFigure}
 { \includegraphics [width=\columnwidth] {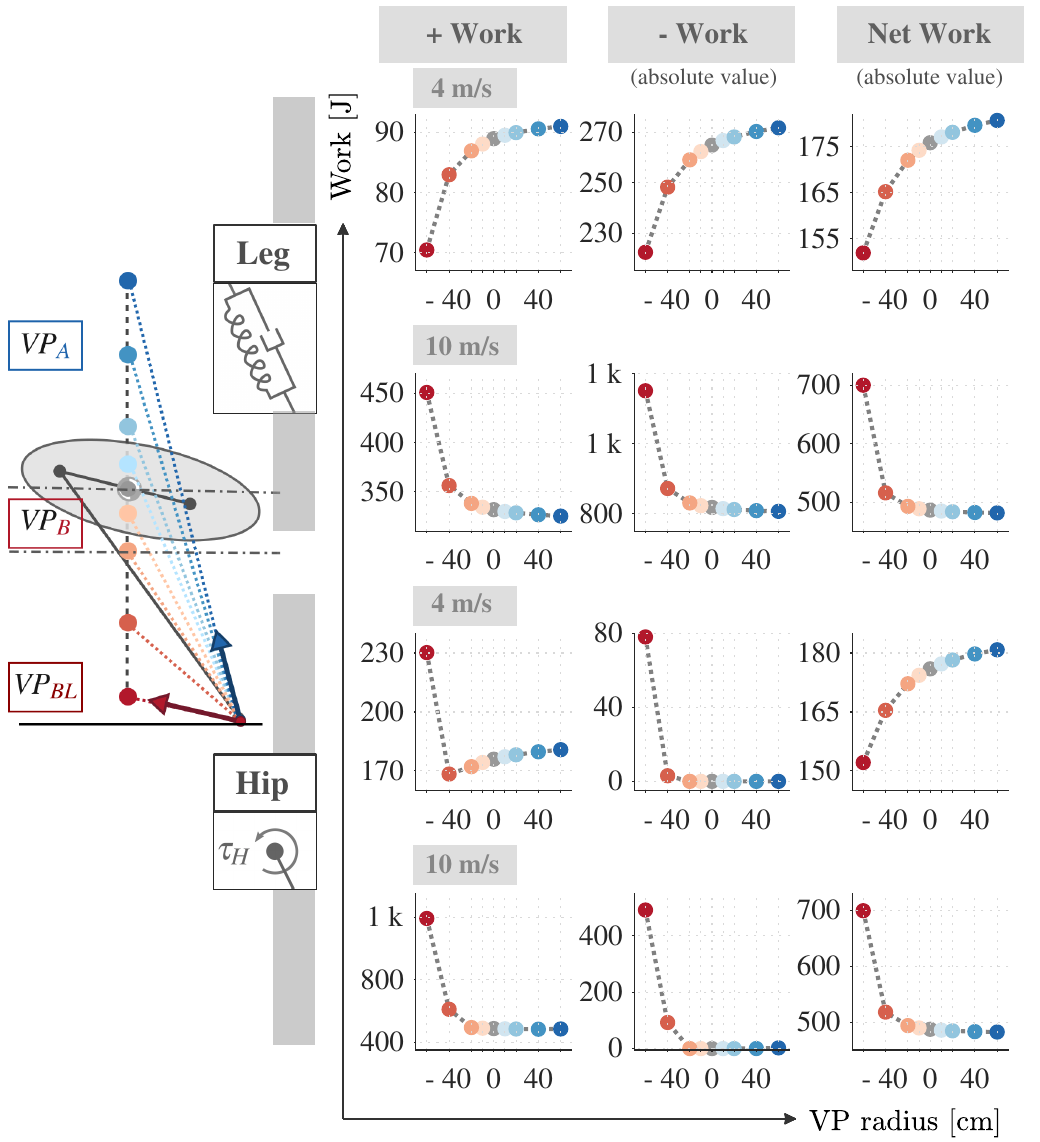}}
      \customSublabelBasic{a)}{0.37,0.925}{color_mygray}
  \customSublabelBasic{b)}{0.57,0.925}{color_mygray}
\customSublabelBasic{c)}{0.79,0.925}{color_mygray}
      \customSublabelBasic{d)}{0.37,0.7}{color_mygray}
  \customSublabelBasic{e)}{0.58,0.7}{color_mygray}
\customSublabelBasic{f)}{0.79,0.7}{color_mygray}
  \customSublabelBasic{g)}{0.37,0.47}{color_mygray}
  \customSublabelBasic{h)}{0.58,0.47}{color_mygray}
\customSublabelBasic{i)}{0.79,0.47}{color_mygray}
  \customSublabelBasic{j)}{0.37,0.245}{color_mygray}
  \customSublabelBasic{k)}{0.58,0.245}{color_mygray}
\customSublabelBasic{l)}{0.79,0.245}{color_mygray}
  \end{annotatedFigure}
\caption{The work performed by the leg (a-f) and hip (g-l) as a function of the \VP location for the speeds of  \SIrange[range-phrase= \,and\, ,range-units=single]{4}{10}{\meter\per\second}. The leg produces net negative work, whereas the hip produces net positive work. At slow speeds, the positive, negative and net work performed by the leg (a-c) increases with a higher \VPa radius and decreases with a higher \VPb radius. The relation is inverse for the fast speeds (d-f). The positive and net hip work displays a relation similar that of the leg  (see g,i and j,l) for \VPa and \VPb with a radius smaller than \SI{30}{\centi\metre} (i.e.,~above the leg axis). For these \VP targets, the hip produces no negative work (h,k). When the \VP radius is set higher than \SI{30}{\centi\metre} , the hip starts to generate negative work (h and k). The excess work is compensated with an increase in the positive hip work to maintain steady state conditions. In other words, the peak work requirement of the hip increases, while the net hip work follows the prior trend.}
\label{fig:WorkDistribution}
\end{figure}

\paragraph{Spatial Analysis: }\label{subsubsecpara:WorkLegHip_rVP} 
We investigate the relation between the \VP location and leg-hip work. We mark that this relation changes with the forward running speed (see \mcref{fig:WorkDistribution}). Therefore, we split our analysis into two parts involving slow and fast speeds of \SIrange[range-phrase=-,range-units=single]{4}{6}{\meter\per\second} and \SIrange[range-phrase=-,range-units=single]{8}{10}{\meter\per\second}. In the following, we derive statements for slow speeds and all effects described are reversed for the fast speeds.
%
We start our analysis with the leg. As the \VP radius increases from \SIrange[range-units=single, range-phrase=\ to\ ]{0}{60}{\centi\meter}; the positive, negative, and net leg work magnitudes (\mcref{fig:WorkDistribution}a) increase 
\SIlist[list-units=single,list-final-separator={, }, list-pair-separator={, }]{3.4; 2.2; 2.2}{\percent}\footnote[8]{\,\% increase and decrease are calculated with reference to the values for $r_{VP}\myeq$\SI{0}{\centi\meter}.} for \VPa ({\protect\markerVPa}), and decrease
\SIlist[list-units=single,list-final-separator={, }, list-pair-separator={, }]{20; 16; 13}{\percent} for \VPb ({\protect\markerVPb}), respectively. 

The hip generates only positive work for \VPa (see \multirefrange{fig:WorkDistribution}{g}{i},\,{\protect\markerVPa}).
The lower the \VPa radius, the less work the hip performs. Hip work increases about \SI{2.2}{\percent} with the increasing \VP radius. 
\VPb up to the radius of \SI{30}{\centi\metre} reduces the work requirement of the hip further about \SI{4}{\percent}~(\mcref{fig:WorkDistribution}g). But, when the \VPb radius is bigger than \SI{30}{\centi\metre} (\VPbl), the hip starts to generate negative work (\multirefrange{fig:WorkDistribution}{g}{h}) and needs to produce large amounts of positive work to compensate for both its own negative work and leg's. For example, at a \VPbl radius of \SI{60}{\centi\meter} the positive hip work requirement increases \SI{30}{\percent} (\mcref{fig:WorkDistribution}g). A larger \VPbl radius yields higher positive and negative hip work independent from the speed. On the other hand, a larger \VPbl reduces the net hip work at slow speeds (\mcref{fig:WorkDistribution}i), which creates a trade-off between the peak torque demand (\mcref{fig:Kinetics}f) and mean energy expenditure. Such a trade-off diminishes at fast speeds of \SI{10}{\metre\per\second}~(see \multirefrange{fig:WorkDistribution}{j}{l}).

When the forward speed increases from 4~to~\SI{10}{\metre\per\second}, the mean values of the spring, damper and hip energies show approximately 2-fold, 3-fold and 2.5-fold increase respectively (see {\protect \markerVPa}{\protect \markerVPb}, \multirefrange{fig:Energy_dx}{d}{f}).

\begin{figure}[t!]
\centering
\begin{annotatedFigure}
 { \includegraphics [width=\columnwidth]{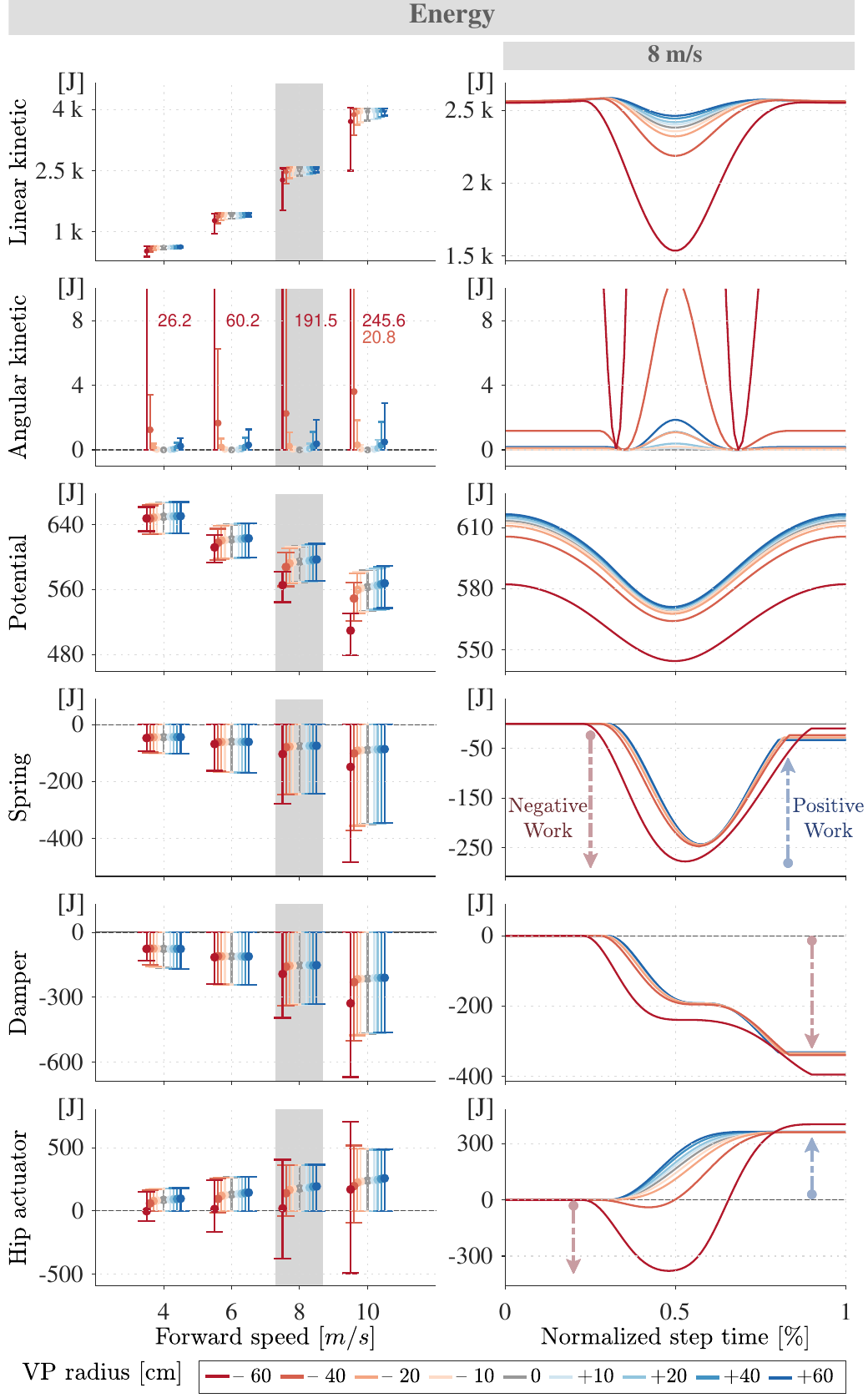}}
  \customSublabelBasic{a)}{0.025,0.94}{color_mygray}
\customSublabelBasic{g)}{0.51,0.94}{color_mygray}
  \customSublabelBasic{b)}{0.025,0.802}{color_mygray}
\customSublabelBasic{h)}{0.51,0.795}{color_mygray}
  \customSublabelBasic{c)}{0.025,0.645}{color_mygray}
\customSublabelBasic{i)}{0.51,0.645}{color_mygray}
  \customSublabelBasic{d)}{0.025,0.5}{color_mygray}
\customSublabelBasic{j)}{0.51,0.5}{color_mygray}
  \customSublabelBasic{e)}{0.025,0.352}{color_mygray}
\customSublabelBasic{k)}{0.51,0.352}{color_mygray}
  \customSublabelBasic{f)}{0.025,0.21}{color_mygray}
\customSublabelBasic{l)}{0.51,0.21}{color_mygray}
\end{annotatedFigure}
\caption{The min.~and max.~values of kinetic~{(a$\text{-}$b)}, potential (c) and external energies (d-f for spring, damper, hip actuator) over forward speed are shown in bar plot, where the mean value is marked with a circle. For the data points that are out of plot boundaries, the max.~values are written numerically. The progression of energies over step time at \SI{8}{\metre\per\second} are given in (g-l). The absolute value of the mean potential energy decreases with speed, whereas the mean linear kinetic, spring, damper and hip energies increase. \VPa yield smaller energy fluctuations for the linear kinetic, potential, spring and damper energies, compared to \VPb. } 
\label{fig:Energy_dx}
\end{figure}

\subsubsection{Energy of the \CoM} \label{subsubsec:EnergyCoM} \hfill 

In this section, we look at how kinetic and potential energies of the \CoM change as a function of forward speed (on the left of \mcref{fig:Energy_dx}) and the normalized step time (on the right). \mcref{fig:Energy_dx}g-\labelcref{fig:Energy_dx}i show that \VPa causes smaller fluctuations in the linear and angular kinetic energies and higher fluctuations in potential energy, compared to \VPb.~We show the energy fluctuations ($\Delta$) that correspond to \multirefrange{fig:Energy_dx}{a}{c} in \multirefrange{fig:Energy_Delta}{a}{c} and observe that \VPa minimizes the linear kinetic and total energy fluctuations of the~\CoM.

When we increase the forward speed from 4 to \SI{10}{\metre\per\second}, the mean linear and angular kinetic energies show approximately 6-fold and 2-fold increase, while the potential energy shows 1.15-fold decrease (see {\protect \markerVPa}{\protect \markerVPb}, \multirefrange{fig:Energy_dx}{a}{c}). In addition, the fluctuations within the distinct types of energies increase, which are indicated with the magnitude of the bar plot in \mcref{fig:Energy_dx}~and are quantified numerically in \mcref{fig:Energy_Delta}. \VPa has lower linear kinetic and higher potential energy fluctuations compared to \VPb. Fluctuations in angular kinetic energy  depend on the \VP radius.

\begin{figure}[!tb]
\centering
\begin{annotatedFigure}
  {\includegraphics [width=\columnwidth]{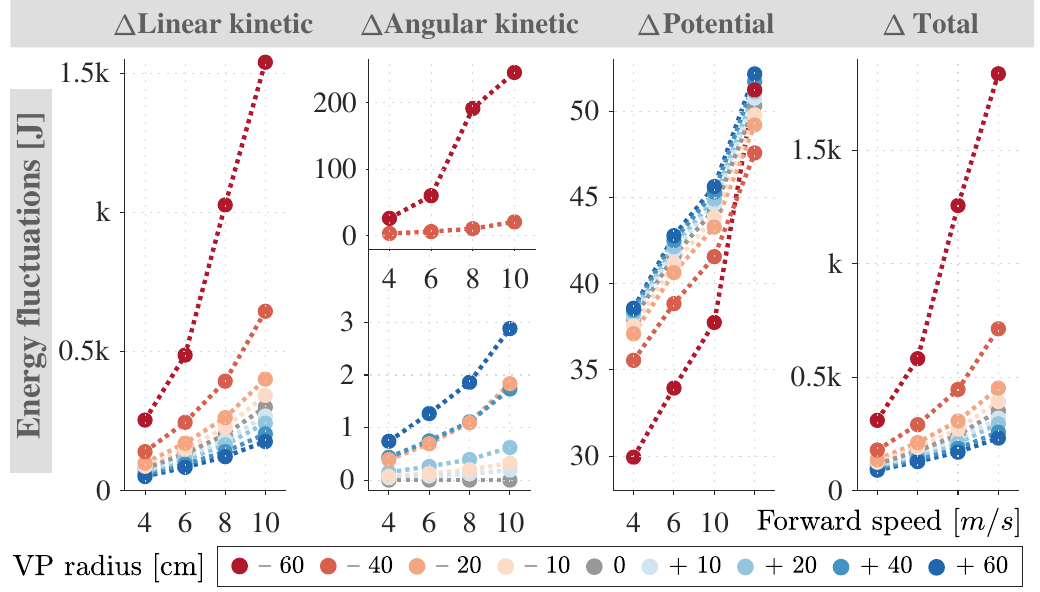}}
  \customSublabelBasic{a)}{0.03,0.885}{color_mygray}
\customSublabelBasic{c)}{0.565,0.885}{color_mygray}
\customSublabelBasic{b.1)}{0.32,0.885}{color_mygray}
\customSublabelBasic{b.2)}{0.32,0.51}{color_mygray}
\customSublabelBasic{d)}{0.79,0.885}{color_mygray}
\end{annotatedFigure}
\caption{The difference ($\Delta$) between min.~and~max. energy over forward speed {\protect \footnotemark}. As the \VP target switches from \VPa to \VPb ({\protect \markerVPb}{\protect \arrowShort}{\protect \markerVPcom}{\protect \arrowShort}{\protect \markerVPa}), the linear kinetic (a) and total (d) energy fluctuations increase, whereas the potential (i.e., vertical) energy fluctuations decrease (c). The angular kinetic energy fluctuations increase with the \VP radius (b.1-b.2). }
\label{fig:Energy_Delta}
\end{figure}
\footnotetext{ \,The bar plot in \multirefrange{fig:Energy_dx}{a}{c} and scatter plot in Figure 16 refer to the same data interpretation.}

\subsubsection{The Mechanical  Cost of Transport}\label{subsec:MCoT} \hfill 

In robotics and biomechanics, the \emph{cost of transport} (CoT) is a measure to compare the energy efficiency of different species and robots \cite{Bhounsule_2012,Tucker_1975}. It is defined as the energy used per traveled distance and expressed as $\mathrm{CoT} \myeq \frac{P}{mg\dot{x}}$ in dimensionless form, where $P$ involves the overall power generated by the metabolism, muscles/actuators and so on. A smaller CoT indicates a better energy economy.

\begin{figure}[!tb]
\centering
\begin{annotatedFigure}
  {\includegraphics [width=\columnwidth] {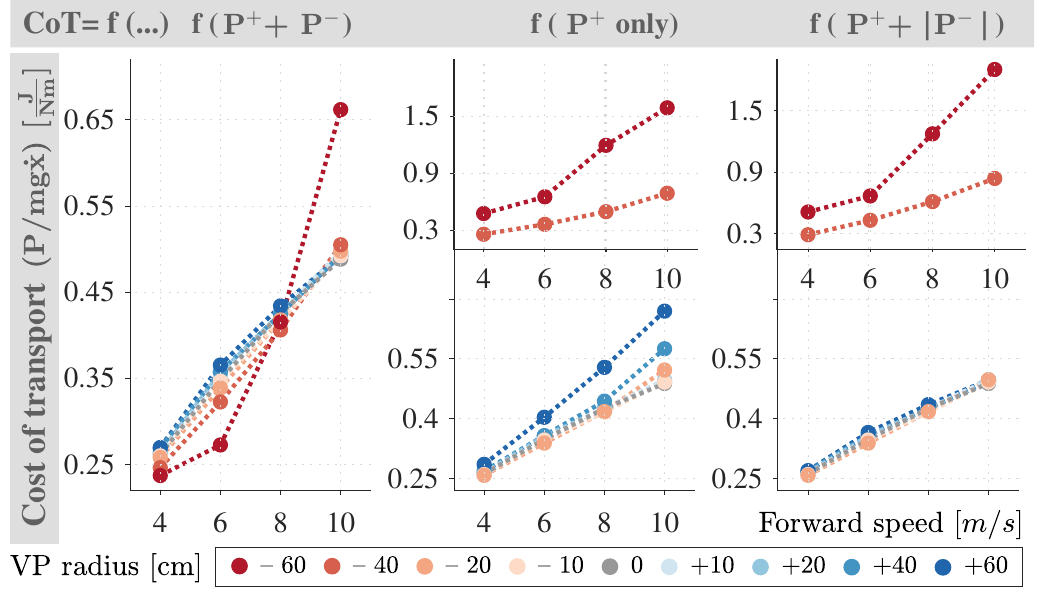}}
    \customSublabelBasic{a)}{0.1,0.87}{color_mygray}
\customSublabelBasic{b.1)}{0.4,0.87}{color_mygray}
\customSublabelBasic{b.2)}{0.4,0.48}{color_mygray}
\customSublabelBasic{c.1)}{0.71,0.87}{color_mygray}
\customSublabelBasic{c.2)}{0.71,0.48}{color_mygray}
\end{annotatedFigure}
\caption{The dimensionless cost of transport (CoT) is calculated using the net (a), positive (b.1-b.2), and absolute (c.1-c.2) values of the hip power. The CoT increases with increasing forward speed, but no trend is visible for changing \VP target.} 
\label{fig:MCoT_RT}
\end{figure}

In our model, the actuator is the hip motor and the CoT reflects hip work. We consider three cases, where we use the (\mcref{fig:MCoT_RT}a) net, (\mcref{fig:MCoT_RT}b) positive, and (\mcref{fig:MCoT_RT}c) absolute values of hip power to calculate the CoT (\mcref{fig:MCoT_RT}). As the running speed increases from 4 to \SI{10}{\metre\per\second}, the CoT increases from 0.25 to 0.65, 1.7 and 1.6 for (a-b-c) respectively. In addition, we see that \VP radii between $\minus$\SI{10}{\centi\metre} and \SI{30}{\centi\metre} yield a small CoT. For higher \VP radii, the CoT increases. We measure the power expenditure in hip joint space. Our measure involves both mechanical and partially metabolic CoT: the hip compensates for the damping in the leg and captures some of the metabolic effects indirectly. 

\begin{table*}[!bht]                                                                                                                                                                                                                                                                                                                                                                                                                                                                                                                                                                 
\floatbox[\capbeside\thisfloatsetup{capbesideposition={right,center},capbesidewidth=4.4cm}]{table}[\FBwidth]
{\caption{Control design rules for avian bipedal locomotion with TSLIP dynamics. These rules also hold for the human morphology, with the exception of R3 \cite{Drama_2019}. The human model shows no speed dependency in energetics. Rules R1-R2 are valid for all speeds.}
\label{tab:DesignRules}}
{\begin{adjustbox}{width=0.7\textwidth}
{\begin{tabular}{@{}  l !{\color{color_VPcom}\vrule} c c c c c  @{}}
\multicolumn{1}{c}{\huge{\raggedright Speed}} & \multicolumn{1}{c}{\huge{Rule}} &   \multicolumn{1}{c}{\huge{Benefit}} &   \multicolumn{1}{c}{\huge{Cost}} & \multicolumn{2}{c}{\huge{Use Case}} \vspace{1mm}  \\ 
\hline \\[\dimexpr-\normalbaselineskip+4pt]
\multirow{6}{*}{\hspace{2mm} \parbox{1cm}{\centering {\Large 4-6} {\large [\si{\meter\per\second}]}}}  &{\centering {\Large \color{color_mygray} R1) } \large \VPb} & \multirow{2}{*}{\parbox{1.5cm}{\centering reduce hip work}}   & \multirow{2}{*}{--}  & \multicolumn{2}{c}{\parbox{7.1cm}{\raggedright 1. Birds with hip extensor strength deficit}} \\ 
  & $r_{VP}\myleq\,$\SI{30}{\centi\meter} &    &    &    \multicolumn{2}{c}{\parbox{7.1cm}{\raggedright 2. Robots with insufficient hip motor limits}} \vspace{1mm}\\
\arrayrulecolor{color_VPcom}\cline{2-6} \\[\dimexpr-\normalbaselineskip+4pt]
  & \multirow{4}{*}{\parbox{2.5cm}{\centering {\Large \color{color_mygray} R2)} { \large \VPbl} $r_{VP}>\,$\SI{30}{\centi\meter}}} & \multirow{4}{*}{\parbox{1.5cm}{\centering reduce leg work}}   & \multirow{4}{*}{\parbox{1.7cm}{\centering high peak hip torque}}  & \multicolumn{2}{c}{\parbox{7.1cm}{\raggedright 1. Birds with reduced leg extensor capabilities}} \\ 
  &  &    &    &    \multicolumn{2}{c}{\parbox{6.1cm}{\raggedright  caused by e.g., knee arthritis or obesity}} \\ 
  &  &    &    &    \multicolumn{2}{c}{\parbox{7.1cm}{\raggedright 2. Robots with weak leg actuation }}\\ 
   &  &    &    &    \multicolumn{2}{c}{\parbox{6.1cm}{\raggedright or light weight legs} }\\
\hline \\[\dimexpr-\normalbaselineskip+4pt]
\multirow{2}{*}{\hspace{2mm} \parbox{1cm}{\centering {\Large 8-10} {\large [\si{\meter\per\second}]}}}  & \multirow{2}{*}{{\Large \color{color_mygray} R3) } \large  \VPa} & \multirow{2}{*}{\parbox{2.5cm}{\centering reduce hip and leg work}}   & \multirow{2}{*}{--}  & \multicolumn{2}{c}{\parbox{7.1cm}{\raggedright 1. Birds that run at high speeds}} \\ 
 &  &    &    &    \multicolumn{2}{c}{\parbox{7.1cm}{\raggedright 2. Robots to increase max. attainable speed}}\\ 
\hline \\[\dimexpr-\normalbaselineskip+4pt]
\multirow{6}{*}{\hspace{2mm} \parbox{1cm}{\centering \Large All Speeds}}  & \multirow{3}{*}{{\Large \color{color_mygray} R4) } \large  \VPa} & \multirow{3}{*}{\parbox{3.8cm}{\centering reduce linear kinetic \& total energy fluctuations of the \CoM}}   & \multirow{3}{*}{\parbox{3cm}{\centering higher angular energy fluctuations of the \CoM}}  & \multicolumn{2}{c}{\parbox{7.1cm}{\raggedright {\color{color_VPcom} \textbullet\ }When stability in horizontal axis is needed }} \\ 
 &  &    &    &    \multicolumn{2}{c}{\parbox{6.1cm}{\raggedright e.g., carrying a fragile load, stepping }} \\
 &  &    &    &    \multicolumn{2}{c}{\parbox{6.1cm}{\raggedright  down, or any change in ground level}}  \\ 
 \arrayrulecolor{color_VPcom}\cline{2-6} \\[\dimexpr-\normalbaselineskip+4pt]
 & \multirow{3}{*}{{\Large \color{color_mygray} R5) } \large \VPb} & \multirow{3}{*}{\parbox{3cm}{\centering reduce potential energy fluctuations of the \CoM}}   & \multirow{3}{*}{\parbox{3cm}{\centering higher angular energy fluctuations of the \CoM}}  & \multicolumn{2}{c}{\parbox{7.1cm}{\raggedright {\color{color_VPcom} \textbullet\ } When stability in vertical axis is needed }} \\ 
 &  &    &    &    \multicolumn{2}{c}{\parbox{6.1cm}{\raggedright e.g., forward perturbations such as }} \\
 &  &    &    &    \multicolumn{2}{c}{\parbox{6.1cm}{\raggedright stumbling} } \vspace{0.5mm} \\ 
\end{tabular}}
\end{adjustbox}}
\vspace{-3mm}
\end{table*}                                                                                                                                                                                                                                                                                                                                                                                                                                                                                                                                                                         

\section{Discussion}\label{sec:Discussion}
In this section, we suggest control design rules for determining the \VP location, and discuss the benefits and drawbacks of each choice. We interpret the~\VP concept and our design rules in relation to the \GRF alignment. Moreover, we explain if and how our simulation results conform to the biomechanical observations regarding the avian trunk oscillations and~\VP.

\subsection{Control Design Rules}\label{subsec:DesignRules}
We suggest the speed dependent design rule in \cref{tab:DesignRules} for avian bipedal locomotion, which  modifies trunk-leg loading through trunk pitch movements. 

At slow running speeds of \SIrange[range-phrase=-,range-units=single]{4}{6}{\metre\per\second}, we suggest using a \VPb with a radius smaller than \SI{30}{\centi\metre} to minimize the hip work. For example, assume a case where we have a robot, and its motion planner outputs a desired hip torque that exceeds the motor limit. We can make this motion feasible by integrating downward trunk oscillations to our control design, which would reduce the peak motor torque demand. Reduced torque demand allows the usage of smaller motors and hence the design of lightweight robots. Alternatively, the strategy could be used in rehabilitation for a patient with weak hip extensor strength to assist the hip. On the other hand, we suggest utilizing a \VPbl to minimize the leg loading, which would also require high peak hip torques. \VPbl can be beneficial for cases where the leg has to be light (e.g., in robot design), or where the leg actuation is weak (e.g., in case of injuries).

At faster running speeds, the system behavior of the pronograde trunk, and hence our disposition of the \VP changes. The relation between the \VP location and leg work reverses after \SI{8}{\metre\per\second}  and the hip after \SI{10}{\metre\per\second}. In this case,  \VPa with a high radius minimizes both the leg and hip work. Consequently, we propose \VPa for energy efficiency at high speeds.

In terms of the \CoM energetics, \VPa minimizes the horizontal and overall energy fluctuations of the \CoM across all speeds. \VPa could be useful for cases where the horizontal accelerations are not desired, for instance when there are changes in the ground level \cite{Vielemeyer_2019}, or when a biped is carrying a fragile load. \VPb can be selected to minimize the vertical energy fluctuations of the \CoM, for instance when the biped stumbles forward and vertical acceleration is not desired for the motion recovery.

\subsubsection{The VP as a method for GRF Alignment }\label{subsec:Human} \hfill 

In this section, we offer a different perspective for interpreting the \VP, where changing \VP location accounts for modifying the ratio between horizontal and vertical \GRF. We refer to this as \emph{GRF manipulation} or \emph{alignment}.

\begin{figure}[!tb]
\centering
 {\includegraphics [width=\columnwidth]{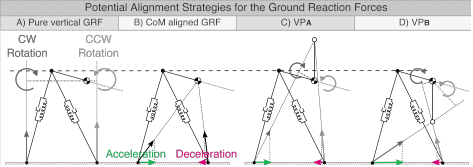}}
\caption{Four potential strategies for the \GRF alignment are illustrated for the foot touch-down and take-off events of the single stance phase of running. The moment created by the \GRF vector around the \CoM is indicated with ({\protect \arrowCircCw},{\protect \arrowCircCcw}), whereas the fore-aft acceleration/deceleration created by the \GRF is indicated with ({\protect \arrowAcc},{\protect \arrowDec}) arrows. As {\protect \mylabelA}{\protect \arrowShort}{\protect \mylabelC}{\protect \arrowShort}{\protect \mylabelB}{\protect\arrowShort}{\protect \mylabelD}, 
the horizontal component of the \GRF increases and the fluctuations in the linear kinetic energy increase. The angular fluctuations of the \CoM increase, as the \GRF points further away from the \CoM.}
\label{fig:ClarkVsVP}
\end{figure}

In biomechanics, the \GRF manipulation is considered as the \emph{running technique} of an animal, where the animal adjusts the forces it applies on the ground. In terms of avian locomotion, two potential \GRF alignment strategies were assessed for quail in \cite{Clark_1975} (see \mcref{fig:ClarkVsVP}). The first strategy has purely vertical \GRF {\,\protect\mylabelA} and the second strategy points the \GRF towards to \CoM {\,\protect\mylabelB}. The first strategy only creates a moment around the \CoM ({\protect \arrowCircCw},{\protect \arrowCircCcw}), which rotates the body in the pitch direction and causes fluctuations in the angular kinetic energy. The second strategy yields no angular motion but decelerates ({\protect \arrowDec}) and accelerates ({\protect \arrowAcc}) the main body successively. In other words, it causes fluctuations in the translational kinetic energy. It is known that quails employ {\protect \mylabelB} in running, despite {\protect \mylabelA} being energetically more economical (i.e., {\protect \mylabelA} resulting in less kinetic energy fluctuations) \cite{Clark_1975}. This preference is motivated by the the excessive pitch angle demand of~{\protect \mylabelA}, which is physically not feasible due to the power limit of the hip actuation.

The \VP concept lies between {\protect \mylabelA} and {\protect \mylabelB}, where the \GRF both induces rotation around the \CoM and acc/decelerates the \CoM in the horizontal direction. In our framework, \GRF vectors become more vertically oriented as \VP location transits from \VPb to \VPa, as {\protect \mylabelD}{\protect \arrowShort}{\protect \mylabelB}{\protect \arrowShort}{\protect \mylabelC}{\protect\arrowShort}{\protect \mylabelA}. \VPa with more vertical \GRF alignment yields smaller fluctuations in the linear and overall \CoM energy, in accordance with \cite{Clark_1975,Usherwood_2012}. 

In the light of this new perspective, we reinterpret our results in \mcref{fig:WorkDistribution} of \mcref{subsubsec:WorkLegHip} for the avian morphology. Both the leg and hip work can be reduced, if the \GRF vectors are oriented more vertically at fast speeds and more horizontally at slow speeds.

\subsection{Gait Measurements vs.\,TSLIP Model}\label{subsec:GaitVsTSLIP}
A set of biomechanical experiments report downward trunk motion during the stance phase of avian running, by using methods such as cineradiography or motion capture \cite{Gatesy_1999,Hancock_2014,Jindrich_2007,Rubenson_2007}. Another independent set of experiments estimate a \VP above the \CoM (\VPa) from the \GRF measurements  \cite{Andrada_2014,Blickhan_2015,Maus_2010}. However, when we implement a \VPa in the \TSLIP model, the simulated gaits display upward trunk motion during the stance phase. This is in conflict with the first set of biomechanical observations (see \cref{tab:BioObv}). In the \TSLIP simulation, a forward trunk motion is obtained when the \VP is set below the \CoM (\VPb).

\begin{table}[b!]                                                                                                                                                                                                                                                                                                                                                                                                                                                                                                                                                                 
\centering                                                                                                                                                                                                                                                                                                                                                                                                                                                                                                                                                                          
\captionsetup{justification=centering}
\caption{Biomechanical observations and TSLIP simulation results for avian running with regard to the \VP location and trunk motion at single stance phase.}
\label{tab:BioObv}
\setlength\extrarowheight{1pt}
\begin{adjustbox}{width=0.97\textwidth}
\begin{tabular}{@{} c c !{\color{color_VPcom}\vrule}  c c c c c @{}}
\multicolumn{2}{c}{Method} & \multicolumn{1}{c}{Trunk Motion} &   \multicolumn{1}{c}{Ref.} &   \multicolumn{1}{c}{\VP Location} & \multicolumn{2}{c}{References}  \\ 
\hline \\[\dimexpr-\normalbaselineskip+4pt]
 \multicolumn{2}{c}{\parbox{2.5cm}{\centering Biomechanical measurements}} &  downward & \cite{Gatesy_1999,Hancock_2014,Jindrich_2007,Rubenson_2007} &  \parbox{2cm}{\centering above \CoM} & \,\! \cite{Andrada_2014,Blickhan_2015,Maus_2010} \vspace{1mm} \\
\arrayrulecolor{color_VPcom}\cline{1-7} \\[\dimexpr-\normalbaselineskip+4pt]
\multicolumn{2}{c}{TSLIP model} &   &  &  &  \\
\arrayrulecolor{color_VPcom}\cline{1-2} \\[\dimexpr-\normalbaselineskip+4pt]
\multicolumn{2}{c}{\VPa} & upward  &  {--} & above \CoM &  {--} \\
\hline \\[\dimexpr-\normalbaselineskip+4pt]
\multicolumn{2}{c}{\VPb} & downward  & {--} & below \CoM &  {--} \\
\end{tabular}
\end{adjustbox}
\end{table}                                                                                                                                                                                                                                                                                                                                                                                                                                                                                                                                                                         

In summary, there is a mismatch between the biomechanical observations and \TSLIP model regarding the coupling between the trunk oscillation direction and \VP location. One possible explanation is related to the disparity between the trunk and whole body dynamics. In human walking, the trunk pitching motion is reported to be in antiphase (\SI{180}{\degree} out of phase) with the whole body pitching motion \cite{Gruben_2012}. Given that humans have relatively heavy limbs \cite{deLeva_1996}, it is plausible that the trunk and whole body dynamics deviate from each other. The \TSLIP model with a \VPa and upward trunk motion might reflect the whole body dynamics, and not necessarily the trunk dynamics \cite{Mueller_2017}. However, this argument is not convincing for the avian species with relatively light legs, where the lower extremities contribute little to the whole body dynamics \cite{Fedak_1982}. The existing research is missing data to provide an antiphase correlation between the trunk and whole body pitch dynamics for human running or any kind of avian gait. 

Another potential explanation is related to the data processing of the \GRF measurements. \GRF signals are noisy, especially at touch-down and take-off events, due to the artifacts that results from the impact, heel-strike and ankle push-off. Consequently, \GRF that belong to the initial and final phases of stance phase are removed when estimating the \VP \cite{Maus_2008_II}.~In addition, \GRF data is filtered to remove noise and drift. The \VP is calculated as the point that minimizes the distances between \GRF vectors. Truncation and modification of \GRF signals might cause an error in estimating the \VP. There is a need for systematic experiments that are tailored to investigate how avians and humans modify the \GRF over a wide range of speeds, and how the trunk and whole body motion fits to this framework. 

The \TSLIP we use in our analysis is a simplified model, which can predict the dynamics of running for the \CoM, close to what we observe in nature. 
The choice of such a simplified model for our analysis creates the question, how well the model predictions conform to more complex models and actual underlying dynamics in real-life cases.
Despite these potential drawbacks, the simplification in \TSLIP enables us to isolate the function of the trunk and leave out the inertial effect of other extremities. We can investigate the effect of trunk motion across various bipedal species with different leg characteristics. 

\section{Conclusion}\label{sec:Conclusion}
In this paper we investigated how trunk pitch oscillations affect the dynamics and energetics of running in terrestrial birds, who possess a horizontal (pronograde) trunk. We used an avian TSLIP model with a virtual point (\VP) control target to create trunk oscillations and analyzed how the \VP location affects the energy economy.

We placed the \VP above (\VPa) and below (\VPb) the center of mass (\CoM), and performed a parameter sweep on the \VP radius to assess its effect on energetics. We showed that a \VPa causes upward and a \VPb causes downward trunk motion during the stance phase.

Our results suggest different potential strategies to place the VP in order to shape the energy distribution. One can use a \VPa to reduce the kinetic energy fluctuations of the~\CoM, while a \VPb reduces the potential energy fluctuations. Both of these come at the expense of increased angular energy fluctuations of the \CoM. At the same time, one can reduce the hip and leg work jointly by using a \VPa at fast speeds of \SIrange[range-phrase=-,range-units=single]{8}{10}{\metre\per\second} and a \VPb at slow speeds of \SIrange[range-phrase=-,range-units=single]{4}{6}{\metre\per\second}. A \VPb below the leg axis (\VPbl) is useful to reduce the leg work further at slow speeds of \SIrange[range-phrase=-,range-units=single]{4}{6}{\metre\per\second} at the expense of the high peak hip torques.

The aim of our study was to show how trunk motion can be leveraged to shape the energy distribution. As future work, we plan to extend our simulation analysis to other bipedal morphologies and test the validity of our control strategies with real robots. 

\section{Acknowledgement}\label{sec:Acknowledgement}
The authors thank the International Max Planck Research School for Intelligent Systems (IMPRS-IS) for supporting {\"O}zge Drama. This work was made possible thanks to a Max Planck Group Leader grant awarded to A.~Badri-Spr{\"o}witz by the Max Planck Society.

\section{ORCHID iDs}\label{sec:Orchid}
\noindent {\"O}zge Drama{\,}\orcidicon{0000-0001-7752-0950} \href{https://orcid.org/0000-0001-7752-0950}{https://orcid.org/0000-0001-7752-0950}

\noindent Alexander Badri-Spr{\"o}witz \orcidicon{0000-0002-3864-7307}  \href{https://orcid.org/0000-0002-3864-7307}{https://orcid.org/0000-0002-3864-7307}

\section*{References}\label{sec:References}
\bibliographystyle{iopart-num}
\bibliography{root}


\end{document}